\documentclass[10pt,twocolumn,letterpaper]{article}

\usepackage[table,xcdraw]{xcolor}
\usepackage{wacv}
\usepackage{times}
\usepackage{epsfig}
\usepackage{graphicx}
\usepackage{amsmath}
\usepackage{amssymb}

\usepackage{mdframed}
\usepackage{makecell}
\usepackage{bm}
\usepackage{float}
\usepackage{graphicx}
\usepackage{listings}
\usepackage{epsfig}
\usepackage{amsmath}
\usepackage{amssymb}
\usepackage{import}
\usepackage{tabularx}
\usepackage{adjustbox}

\wacvfinalcopy % *** Uncomment this line for the final submission

\def\wacvPaperID{1072} % *** Enter the wacv Paper ID here

\ifwacvfinal\fi
\begin{document}

%%%%%%%%% TITLE
% \title{Differentiable Approximation Bridges For Training Networks Containing Non-Differentiable Functions}
\title{Improving Discrete Latent Representations \\With Differentiable Approximation Bridges}

\newcommand*\samethanks[1][\value{footnote}]{\footnotemark[#1]}

% Authors at the same institution
%\author{First Author \hspace{2cm} Second Author \\
% Authors at different institutions
% \author{First Author \\
% \and
% Second Author \\

% \author{Jason Ramapuram \\
% University of Geneva \&\\ University of Applied Sciences,
% {\tt\small jason@ramapuram.net}
% }
% \and
\author{Jason Ramapuram \thanks{Equal Contribution}\ \ \thanks{Work
    done during Apple internship.} \\
  University of Geneva \&\\ University of Applied Sciences. \\
  Geneva, Switzerland \\
  {\tt\small jason@ramapuram.net}
  \and
  Russ Webb \samethanks[1]\\
  Apple Inc. \\
  One Apple Park Way \\
  Cupertino, CA 95014 \\
  {\tt\small rwebb@apple.com}
}

% \author{
% -  Jason Ramapuram\thanks{Equal Contribution}\ \ \thanks{Work done during Apple internship.}\\
% -  University of Geneva \&\\ University of Applied Sciences, \\
% -  Switzerland \\
% -  \texttt{jason@ramapuram.net}
% -  \And
% -  Russ Webb\samethanks[1] \\
% -  Apple Inc. \\
% -  One Apple Park Way \\
% -  Cupertino, CA 95014 \\
% -  \texttt{rwebb@apple.com}
% -}

\maketitle
\ifwacvfinal\thispagestyle{empty}\fi

%%%%%%%%% ABSTRACT
\begin{abstract}
Modern neural network training relies on piece-wise
(sub-)differentiable functions in order to use backpropagation to update model parameters. In this work, we introduce a
novel method to allow simple non-differentiable functions at intermediary
layers of deep neural networks. We do so by training with a
differentiable approximation bridge (DAB) neural network which
approximates the non-differentiable forward function and provides
gradient updates during backpropagation. We present strong empirical
results (performing over 600 experiments) in four different domains:
unsupervised (image) representation learning, variational (image)
density estimation, image classification, and sequence sorting to
demonstrate that our proposed method improves state of the art
performance. We demonstrate that training with DAB aided
discrete non-differentiable functions improves image
reconstruction quality and posterior linear separability by
\textbf{10\%} against the Gumbel-Softmax relaxed estimator
\cite{maddison2016concrete,jang2016categorical} as well as providing a \textbf{9\%}
improvement in the test variational lower bound in comparison to the
state of the art RELAX \cite{grathwohl2017backpropagation} discrete estimator. We
also observe an accuracy improvement of \textbf{77\%} in neural
sequence sorting and a \textbf{25\%} improvement against the
straight-through estimator \cite{bengio2013estimating} in an image
classification setting. The DAB network is not used for inference and expands the class of functions that
are usable in neural networks.
\end{abstract}

% keywords can be removed
% \keywords{Non-Differentiable Functions \and Neural Networks \and
%   Gradient Approximators}

%   -------------------------------------------------------------------------

% \vspace{-0.2in}
\section{Introduction}

% Deep neural networks have advanced the state of the art in
% object recognition \cite{he2016deep,szegedy2016inception}, machine
% translation \cite{devlin2019bert} and game playing
% \cite{silver2016mastering}, however they have trouble
% generalizing outside of the range of numerical values encountered
% during training \cite{trask2018neural}. In contrast, traditional (non-learned)
% algorithms such as Merge-Sort \cite{katajainen1997meticulous}, do not
% suffer from this problem. In this work, we aim to bridge the
% gap between these two paradigms by allowing for
% non-differentiable functions, such as Merge-Sort, to be used in
% neural network pipelines.

Deep neural networks have advanced the state of the art in
object recognition \cite{he2016deep,szegedy2016inception}, machine
translation \cite{devlin2019bert}, and game playing
\cite{silver2016mastering}, however they generally only function over the range of numerical values encountered
during training \cite{trask2018neural}. In contrast, traditional (non-learned) algorithms, such as Merge-Sort
\cite{katajainen1997meticulous}, are provably stable and can deal
with arbitrary inputs. In this work, we introduce a novel formulation
that allows for the incorporation of simple non-differentiable
functions, such as Merge-Sort, Signum and K-Means in neural network pipelines.

Most state of the art neural networks
\cite{he2016deep,szegedy2016inception,goodfellow2014generative} rely
on some variant of Robbins-Monroe \cite{robbins1951stochastic} based
stochastic optimization. The requirement for utilizing this algorithm
includes the assumption that the gradients of the functional
be Lipschitz continuous. In contrast, some of the most common functions used in neural networks, the ReLU
activation \cite{agarap2018deep} and the Max-Pooling layer
\cite{wu2015max} are not fully differentiable. In general,
this problem is circumvented by ignoring the measure zero
non-differentiable domain or through the use of the adjoint
method. Functions such as \emph{sort} and \emph{k-means} are not amenable to a similar treatment.

% In addition, discrete latent variables are found throughout machine
% learning, from reinforcement learning to gaussian mixture
% models. However,

In this work, we study approximate gradient pathways that allow for
simple non-differentiable functions as sub-modules of neural
networks. We validate DAB using the \emph{sort}, \emph{top-k},
\emph{k-means}, \emph{signum}, \emph{binary-threshold} and
\emph{non-reparameterized bernoulli} non-differentiable functions and
demonstrate competitive performance on a variety of tasks. DAB
enables the use of these functions by introducing a smooth neural network
approximation to the non-differentiable function; the gradients of the
DAB network are then used at training time to update previous layers
of the network. The DAB network is trained jointly with the central optimization objective and creates
its approximation via the introduction of a regularizer (Section
\ref{model_sec}). At inference, the DAB network is removed, thus
requiring no extra memory or compute after training.

% One of the most commonly used activations, the ReLU activation
% \cite{agarap2018deep} function is not fully differentiable, however
% this problem is simply overcome by ignoring the non-differentiable
% point at zero.

%%% Local Variables:
%%% mode: latex
%%% TeX-master: "main"
%%% End:

% \vspace{-0.1in}
\section{Related Work}
% \vskip -0.2in
\begin{table*}
  \scalebox{0.65}{\parbox{1.0\linewidth}{%
      {\renewcommand{\arraystretch}{1.3}%

\begin{tabular}{l|l|c|c|c|c|}
\cline{2-6}
                                                          &
                                                            \multicolumn{1}{c|}{\textbf{Method
                                                            /
                                                            Objective}}
  & \textbf{\begin{tabular}[c]{@{}c@{}}Supports \\ Non-Differentiable
              \\ Functions\end{tabular}} &
                                           \textbf{\begin{tabular}[c]{@{}c@{}}Scales to \\ Large \\ Dimensions\end{tabular}} & \textbf{\begin{tabular}[c]{@{}c@{}}Works with\\ Operators that\\ Change Dimension\end{tabular}} & \textbf{\begin{tabular}[c]{@{}c@{}}Typical\\Unique Hyper\\ Parameters \end{tabular}}  \\ \hline
\multicolumn{1}{|l|}{\textbf{DNI \cite{jaderberg2017decoupled} /
  DPG \cite{huo2018training} / DGL \cite{belilovsky2019decoupled}}}
                                                          &
                                                            Asynchronous
                                                            network
                                                            updates.
  & no
                                         & \textbf{yes}
                                                                                                                             & \textbf{yes} & -                                                                                             \\ \hline
\multicolumn{1}{|l|}{\textbf{Gradient Free Methods
  \cite{van1987simulated, kennedy2010particle, goldberg1988genetic,
  asselmeyer1997evolutionary, gelfand1990sampling, gilks1995markov,
  lillicrap2016random}}}      & Optimize arbitrary functions.
  & \textbf{yes}
                                         & no
                                                                                                                             & \textbf{yes} & -                                                                                            \\ \hline
\multicolumn{1}{|l|}{\textbf{Score Function Estimator
  \cite{glynn1990likelihood,kleijnen1996optimization}}}   &
                                                            Differentiate non-differentiable functions. & \textbf{yes}                                                                                          & no                                                                                & \textbf{yes} & \textbf{0}                                                                                            \\ \hline
\multicolumn{1}{|l|}{\textbf{Straight-Through Estimator
  \cite{bengio2013estimating}}} & Ignore non-differentiable
                                  functions. & \textbf{yes}
                                         & \textbf{yes}
                                                                                                                             & no & \textbf{0}                                                                                             \\ \hline
\multicolumn{1}{|l|}{\textbf{Relaxed Estimators
  \cite{maddison2016concrete,jang2016categorical, tucker2017rebar, grathwohl2017backpropagation} }}                 &
                                                            Relaxed
                                                                                                                      approximations
                                                                                                                      to
                                                                                                                      non-differentiable
                                                                                                                      functions. &
                                                                                                                          \textbf{yes}
                                         & \textbf{yes}
                                                                                                                             &
                                                                                                                               \textbf{yes}
                                                                                                                                                                                                                               &
                                                                                                                                                                                                                                 1 - 3                                                                                             \\ \hline
\multicolumn{1}{|l|}{\textbf{DAB (ours)}}                 &
                                                            Differentiate
                                                            non-differentiable
                                                            functions. &
                                                                         \textbf{yes}
                                         & \textbf{yes}
                                                                                                                             &
                                                                                                                               \textbf{yes}
                                                                                                                                                                                                                               & 1                                                                                             \\ \hline
\end{tabular}
}}}
\vspace{-0.1in}
\end{table*}

\noindent \textbf{Traditional Solutions}: Traditional solutions to handling
non-differentiable functions in machine learning include using the score function estimator (SFE)
\cite{glynn1990likelihood,kleijnen1996optimization} (also known as
REINFORCE \cite{williams1992simple}), the straight-through estimator
(STE) \cite{bengio2013estimating}, or the reparameterization path-wise estimator
\cite{kingma2014}. While the SFE is an unbiased
estimate of the gradients, it generally suffers from high variance
\cite{grathwohl2017backpropagation} and needs to be augmented with
control variates \cite{glasserman2013monte} that require manual tuning
and domain knowledge. The STE on the other hand is a
solution that simply copies gradients back,
skipping the non-differentiable portion (i.e. treating it as an
identity operation). Furthermore, the STE does not allow for operators
that change dimension, i.e. $f: \mathbb{R}^{A} \mapsto \mathbb{R}^B, A
\neq B$, since it is unclear how the gradients of the
larger/smaller output would be copied back. In contrast to the SFE,
the reparameterization trick used in variational autoencoders (VAE) \cite{kingma2014}, enables differentiating through distributions by
reframing the expectation with respect to a variable that is not part
of the computational graph. The
difference between the SFE and the reparameterization trick can be
understood by analyzing how they estimate gradients:

{\centering
  \begin{table}[H]
    \begin{center}
{\renewcommand{\arraystretch}{1.3}%
\begin{tabular}{|l|}
\hline
\textbf{SFE}: {$\!\begin{aligned}
   \nabla_{\theta}\mathbb{E}_{q_{\theta}} [ f(z) ] = \mathbb{E}_{q_{\theta}}[f(z)
   \nabla_{\theta}\log q_{\theta}(z)]
 \end{aligned}$}  \\ \hline
  \textbf{Reparameterization}: {$\!\begin{aligned}
   \nabla_{\theta}\mathbb{E}_{q_{\theta}} [f(z)] = \mathbb{E}_{p(\epsilon)}[\nabla_{\theta}f(z)]
 \end{aligned}$} \\ \hline
\end{tabular}
}
\end{center}
\end{table}
}
\vspace{-0.2in}
\noindent The reparameterization path-wise estimator takes into account how the
derivative of the function, $f(z)$, is modulated by the choice of
measure, $q_{\theta}$, while the SFE treats the function, $f(z)$, as a
black box. See \cite{DBLP:journals/corr/abs-1906-10652} for
a thorough treatment on the topic.\\

\noindent \textbf{Relaxed Differentiability}: While the
reparameterization estimator has lower variance than the SFE, it did
not afford a discrete reparameterization until recently. The
Gumbel-Softmax relaxed estimator (simultaneously proposed in
\cite{maddison2016concrete,jang2016categorical}) for the Bernouilli
and Discrete distributions anneals a softmax with additive Gumbel
noise until it eventually converges to the corresponding `hard'
distribution. This technique can be interpreted as a form of bias-variance
trade-off \cite{maddison2016concrete}. More recently, REBAR
\cite{tucker2017rebar} and RELAX \cite{grathwohl2017backpropagation}
were proposed to combine both the SFE and the reparameterization trick
in order to produce a new estimator with lower variance. RELAX
\cite{grathwohl2017backpropagation} differs from REBAR
\cite{tucker2017rebar} by using a learned Q function
\cite{watkins1992q} neural network as a control variate. % In contrast
% to these estimators, our solution does not require extensive parameter
% tuning.
We empirically contrast our estimator against
Gumbel-Softmax in Experiment \ref{unsup} and RELAX and REBAR in
Experiment \ref{other_estimators_exp}, demonstrating improved
performance using three different metrics.\\

\noindent \textbf{Gradient Free Methods}: Machine learning has a
rich history of backpropagation alternatives, ranging from simulated
annealing \cite{van1987simulated}, particle swarm optimization
\cite{kennedy2010particle}, genetic algorithms
\cite{goldberg1988genetic}, evolutionary strategies \cite{asselmeyer1997evolutionary}, and
Bayesian approaches such as MCMC based sampling algorithms
\cite{gelfand1990sampling, gilks1995markov}. These algorithms have
been shown \cite{rios2013derivative} to not scale to complex, large
dimension optimization problems in large neural network models.
More recent work in the analysis of backpropagation alternatives
\cite{lillicrap2016random} has demonstrated the possibility of
learning weight updates through the use of random matrices; however, no
statement was made about training / convergence time.\\

\begin{figure*}
  \begin{minipage}{0.33\linewidth}
    \includegraphics[width=\linewidth]{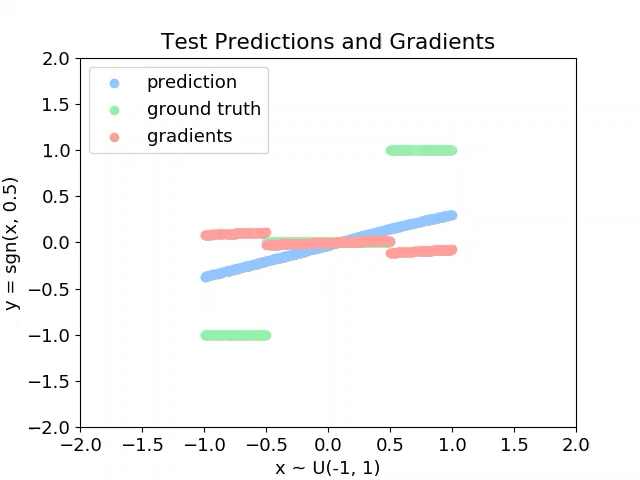}
  \end{minipage}%
  \begin{minipage}{0.33\linewidth}
    \includegraphics[width=\linewidth]{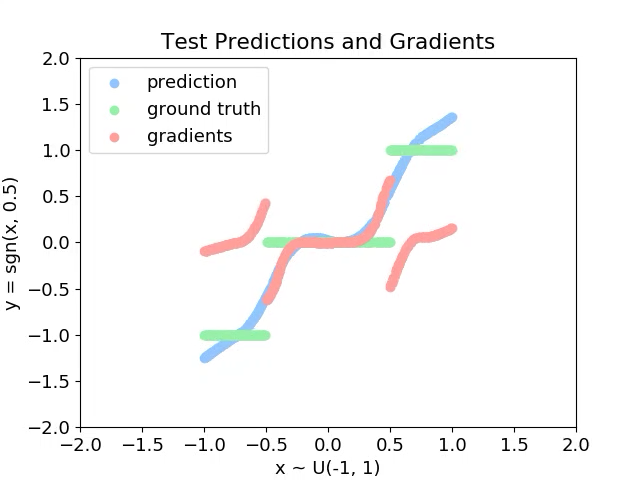}
  \end{minipage}%
  \begin{minipage}{0.33\linewidth}
    \includegraphics[width=\linewidth]{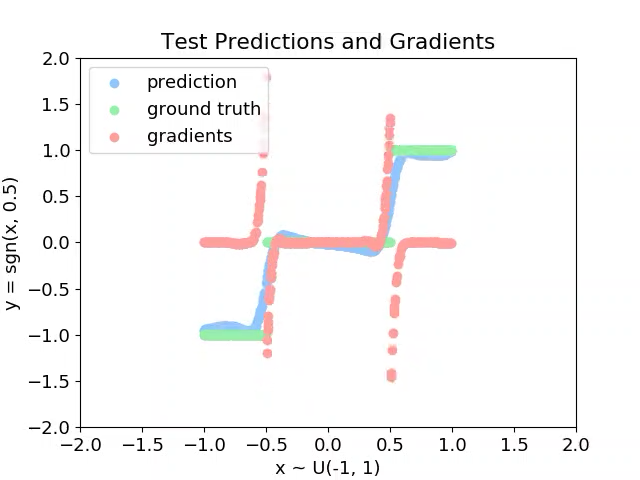}
  \end{minipage}
  \caption{Demonstration of a simple DAB approximation: scatter plots visualize test predictions and gradients of a multi-layer
    dense network fitting an $\epsilon$-margin signum function
    ($\epsilon = 0.5$) over an input space of
    U(-1, 1). \emph{Left}: Near beginning of training. \emph{Middle}:
    Mid-way through training. \emph{Right}: End of training.}\label{simple_nn}
  \end{figure*}%

\noindent \textbf{Asynchronous Neural Network Updates}:
% \cite{jaderberg2017decoupled} characterizes three main bottlenecks
% associated with the updating of neural networks: forward, update \& backward unlocking. Backward unlocking
% allows simultaneous gradient estimation / calculation of all layers in
% a network. Update locking in turn allows for the updating of all
% relevant model parameters in a simultenous way. Finally, forward
% unlocking would allow all layers to be able to independently update
% their inferences.
Recent work such as decoupled neural interfaces
(DNI) \cite{jaderberg2017decoupled} and decoupled parallel
backpropagation (DPG) \cite{huo2018training} introduced an auxiliary network
to approximate gradients in RNN models.  Similar approximation
techniques have been introduced \cite{belilovsky2019decoupled} (DGL) to
allow for greedy layerwise CNN based training. The central objective with these models is to enable
asynchronous updates to speed up training time. Our work differs from
all of these solutions in that our
objective is not to improve training speed / parallelism, but to learn
a function approximator of a non-differentiable function such that it
provides a meaningful training signal for the preceding layers in the
network. This approach allows us to utilize non-differentiable
functions such as \emph{kmeans}, \emph{sort}, \emph{signum}, etc, as intermediary
layers in neural network pipelines.

%%% Local Variables:
%%% mode: latex
%%% TeX-master: "main"
%%% End:

% \input{Preliminaries}
\section{Model}\label{model_sec}

% Given a training and test dataset,
% $\mathcal{D}_{tr} = (X_{tr}=\{x_i\}_{i=1}^N, Y_{tr}=\{y_i\}_{i=1}^N)$ and
% $\mathcal{D}_{te} = (X_{te}=\{x_i\}_{i=1}^N, Y_{te}=\{y_i\}_{i=1}^N)$,  we seek
% to train to a model, $f_{\theta}: x_i \mapsto y_i$, using the training
% dataset $\mathcal{D}_{tr}$, such that it generalizes well on a
% predefined metric $\mathcal{M}(f_{\theta}(X_{te}), Y_{te})$, evaluated
% on the hold out test dataset, $\mathcal{D}_{te}$.

\begin{figure}[H]
  \begin{center}
    \includegraphics[width=85mm]{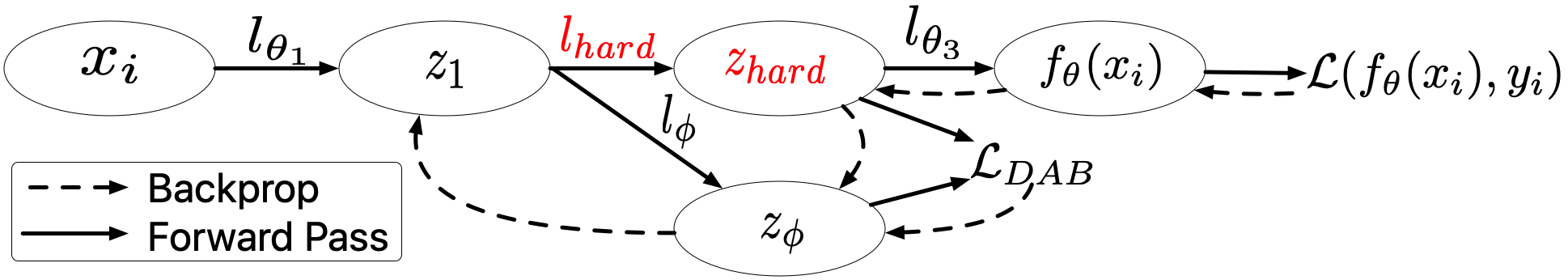}
\end{center}
\caption{Model of our proposed framework. $\color{red}l_{hard}$ represents
  the non-differentiable function and $\color{red}z_{hard}$ its outputs.}\label{graphical_model_v2}
\end{figure}%

Given a set of training input images, $X_{tr} = \{x_i\}_{i=0}^N,\ x_i \in
\mathbb{R}^{M \times M}$, coupled with related ground-truth labels,
 $Y_{tr} = \{y_i\}_{i=0}^N,\ y_i \in \mathbb{R}^J \text{(for
classification) or } y_i \in \mathbb{R}^{M \times M} \text{(for
autoencoding)}$, our objective is to learn the parameters, $\theta$, of
a model, $f_{\theta}: x_i \mapsto y_i$, such that we generalize
well on a hold-out test set, $\{X_{te}, Y_{te}\}$, evaluated with a
predefined metric, $\mathcal{M}(f_{\theta}(X_{te}),  Y_{te})$. We focus on the case where
the parameters $\theta$ are optimized using stochastic gradient
descent \cite{robbins1951stochastic}, operating
on a differentiable loss function $\mathcal{L}(f_{\theta}(X_{tr}), Y_{tr})$.
% Ideally the evaluation metric, $\mathcal{M}$, should be equivalent to the optimization
% objective, $\mathcal{L}$; this is however not always the
% case. Commonly used evaluation metrics such as accuracy,
% precision, recall (used in classification) as well as perceptual metrics such as MS-SSIM
% \cite{wang2003multiscale} and FID \cite{heusel2017gans} (used for
% evaluating image reconstructions) are not differentiable and are thus
% not typically used during optimization.

% In the case of autoencoding the metric $\mathcal{M}$ is equivalent to the
% optimization objective, $\mathcal{L}(f_{\theta}(X_{tr}),
% Y_{tr})$, however in classification common metrics are accuracy,
% precision and recall \footnote{These metrics are not directly
%   differentiable and are thus not used during optimization.}.

% to a target distribution,
% $p_{\theta}(y | x^*)$, by optimizing the parameters, $\theta$, of the model. In the case of
% classification, this output distribution is typically a Categorical
% over the number of target classes. Given a regression or
% auto-encoding task, the distribution is typically modeled as a Gaussian \footnote{We provide a fully Bayesian
% treatment of DAB framework for the curious reader in Appendix
% \ref{bayesian_model_sec}}.
We begin by decomposing our model, $f_{\theta}$, into a layer-wise
representation, $f_{\theta} = l_{\theta_3} \circ l_{\theta_2} \circ
l_{\theta_1}$. We represent the output of each of the above
layers as $\{z_3, z_2, z_1\}$ respectively.
In this work we explore replacing the differentiable
function $l_{\theta_2}$ with a non-differentiable version,
${\color{red} l_{hard}}$. Directly swapping ${\color{red} l_{hard}}$ for $l_{\theta_2}$ is not
viable since it would prevent backpropagation gradient updates of
$l_{\theta_1}$.

In order to circumvent this, we draw inspiration by visualizing what a
typical multi-layer dense neural network does when it fits an output
that is discontinuous. In Figure \ref{simple_nn} we visualize three (test) snapshots of a network
during training. The \emph{ground truth}
target, $y_i$, is an $\epsilon$-margin signum function, $y_i = sgn(x_i,
\epsilon)$, where $x_i \sim U(-1, 1)$ and $\epsilon = 0.5$.

\vspace{-0.1in}
\begin{align}
  sgn(x_i, \epsilon) &= \begin{cases}
      -1 & x_i < -\epsilon \\
      0 & x_i \in [-\epsilon, \epsilon] \\
      1 & x_i > \epsilon
   \end{cases} \label{signum_fn}
\end{align}

\noindent We train a small three layer ELU dense network
using Adam \cite{kingma2014adam} and use the final layer's (un-activated) output in a least-square loss, $\mathcal{L} = ||
y_i - f_{\theta}(x_i) || ^2_2$.
Neural networks have been extensively used to interpolate latent space
in models such as variational autoencoders \cite{kingma2014} and
Word2Vec \cite{mikolov2013efficient}. These interpolations are possible because neural
networks learn a smooth K-Lipschitz mapping to transform the inputs to
their target outputs, which implies smooth gradients. We rely on this property and
introduce the DAB network, $l_{\phi}$.

The DAB network receives $z_1$ from the
previous layer, $z_1 = l_{\theta_1}(x)$, and produces an output, $z_{\phi} = l_{\phi}(z_1)$. This output is constrained to be close to the output of the
hard function, ${\color{red} z_{hard}} = {\color{red}l_{hard}}(z_1)$,
through the introduction of an L2 regularizer  \footnote{An analysis of the choice of
regularizer and convergence is provided in Appendix Section \ref{regularizer}.},
$\mathcal{L}_{DAB} = \gamma || {\color{red}z_{hard}} - z_{\phi} ||_2^2$, where
$\gamma$ represents represents a hyper-parameter that controls the
strength of the regularization. We observed that the choice of
$\gamma$ did not have a strong effect and thus did not conduct an
extensive hyper-parameter search to optimize it. Our final
optimization objective is our typical objective,
% $p_{\theta}(y | x^*)$, coupled with the regularizer described above:
$\mathcal{L}(f_{\theta}(x_i), y_i)$ and the regularizer, $\mathcal{L}_{DAB}$, described above:

% \begin{align}
%   \min_{\theta, \phi} \ \ \ \  \underbrace{p_{\theta}(y | x^*)}_{\text{typical loss}} +
%   \underbrace{\gamma || z_{hard} - z_{\phi} ||_2^2}_{\text{dab loss}}
%   \end{align}
\begin{align}
  % \min_{\theta, \phi} \ \ \ \  \underbrace{\mathcal{L}(f_{\theta}(x), y)}_{\text{typical loss}} +
  % \underbrace{\gamma || z_{hard} - z_{\phi} ||_2^2}_{\text{dab loss}} \\
  % \mathcal{L}(f_{\theta}(x), y) =
  % \mathcal{L}(l_{\theta_3}(l_{hard}(l_{\theta_1}(x)))) \\
  % || z_{hard} - z_{\phi} ||_2^2 = || l_{hard}(z_1) - l_{\phi}(z_1)
  % ||_2^2
  \min_{\theta, \phi}
  \underbrace{\mathcal{L}(l_{\theta_3}({\color{red} l_{hard}}(l_{\theta_1}(x_i))), y_i)}_{\text{typical
  loss}} + \underbrace{\gamma ||
    {\color{red} l_{hard}}(z_1) - l_{\phi}(z_1)
  ||_2^2}_{\text{DAB
  loss}(\mathcal{L}_{DAB})}\label{dab_loss}
\end{align}

The main difference between a typical model and the DAB aided model
presented above is that we use ${\color{red} l_{hard}}$ during the
forward functional evaluations at both training and test time. During training, the model
returns the DAB model's smooth K-Lipschitz gradients, $\frac{\delta
  l_{\phi}}{\delta l_{\theta_1}}$, to update the parameters
$\theta_1$, thus allowing the entire model to be trained
end-to-end. At inference, the DAB model is \emph{completely discarded,
requiring no extra memory or compute}.

%%% Local Variables:
%%% mode: latex
%%% TeX-master: "main"
%%% End:

\begin{figure*}
\begin{minipage}[t]{\textwidth}
\begin{figure}[H]
  \begin{center}
    \includegraphics[width=\linewidth]{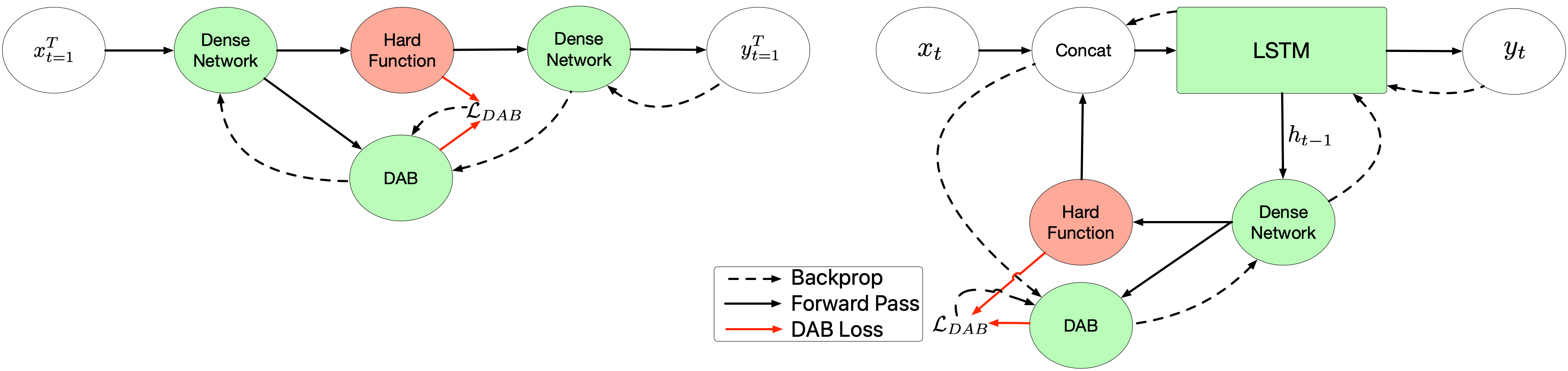}
\end{center}
\caption{\emph{Left}: Dense sorting model with
  non-differentiable-function. \emph{Right}: LSTM Model with a non-differentiable function.}\label{model}
\end{figure}%
\end{minipage}%
\end{figure*}
{\centering
  \begin{table*}
    \begin{center}
  \scalebox{0.9}{\parbox{1.0\linewidth}{%
      {\renewcommand{\arraystretch}{1.3}%
\begin{tabular}{l|l|l|l|l|l|}
\cline{2-6}
Length (T)                 & \textbf{ELU-Dense} &
                                                  \textbf{Ptr-Net\cite{vinyals2015pointer}} & \multicolumn{1}{c|}{\textbf{\begin{tabular}[c]{@{}c@{}}Read-Process\\ Write\cite{vinyals2015order}\end{tabular}}} & \textbf{Signum-RNN (ours)}                                             & \textbf{Signum-Dense (ours)} \\ \hline
\multicolumn{1}{|l|}{T=5} & 86.46 $\pm$ 4.7\% (x5) & 90\% & 94\% &
\textbf{99.3 $\pm$ 0.09\% (x5)} & \textbf{99.3 $\pm$ 0.25\% (x5)} \\ \hline
\multicolumn{1}{|l|}{T=10} & 0 $\pm$ 0\% (x5) & 28\% &
57\% & 92.4 $\pm$ 0.36\% (x5) &
\textbf{94.2 $\pm$
0.1\% (x5)} \\ \hline \multicolumn{1}{|l|}{T=15} & 0
$\pm$ 0\% (x5) & 4\% & 10\% & \textbf{87.2
$\pm$ 0.3\% (x5)} & 79.8 $\pm$ 0.8\% (x5) \\ \hline
\end{tabular}}
}}
\end{center}
\caption{All-or-none sorting test-accuracy (presented as
  mean $\pm$ std (replication)) for varying length (T) sequences.}\label{sort_results}
\end{table*}
}

\begin{figure*}
  \begin{minipage}{0.5\linewidth}
    \includegraphics[width=\linewidth]{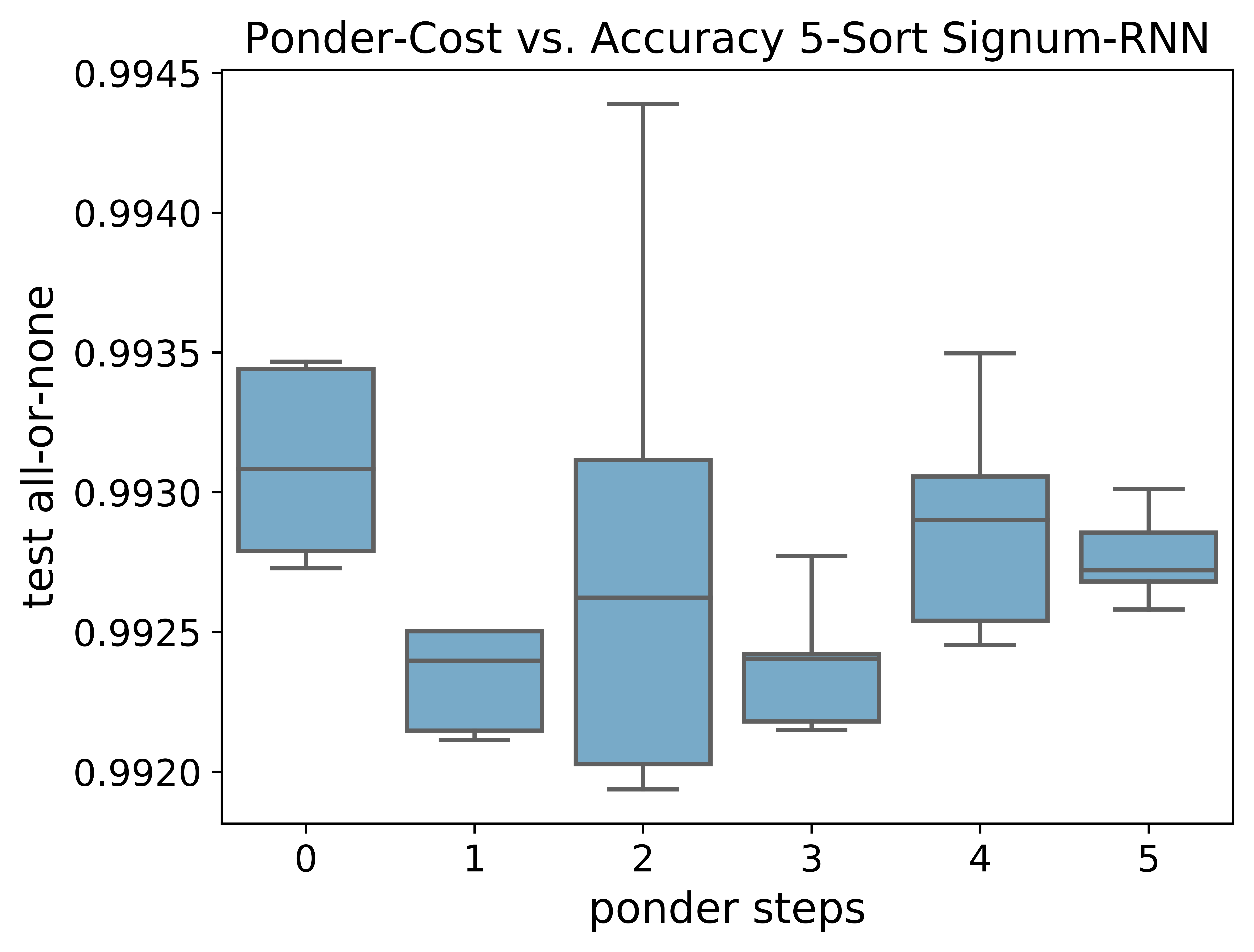}
  %   \includegraphics[width=\linewidth]{./imgs/ponder.png}
  %   \caption{Effect of increasing ponder duration for 5-sort problem. Each
  % ponder loop was run 5 times.}\label{ponder}
  \end{minipage}%
  \begin{minipage}{0.5\linewidth}
    \includegraphics[width=\linewidth]{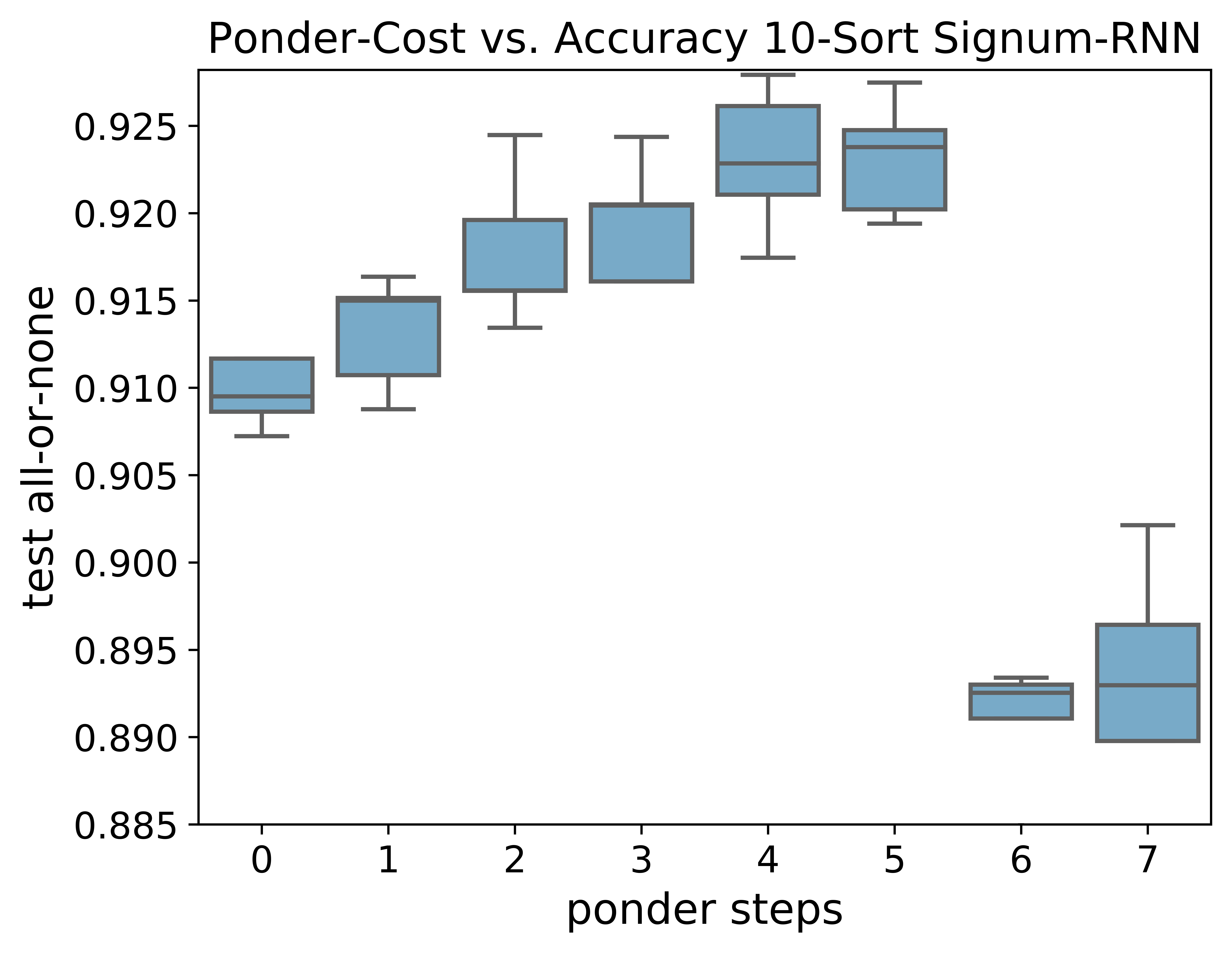}
  \end{minipage}
  \caption{Effect of increasing ponder steps for 5-sort (\emph{left}) and
    10-sort (\emph{right}) problems. The mean and standard deviation of the maximum test all-or-none accuracy are
    reported over 5 trials per ponder length.}\label{ponder}
  \vskip -0.2in
  \end{figure*}%

\section{Experiments}

We quantify our proposed algorithm on four different benchmarks:
sequence sorting, unsupervised (image) representation learning,
variational (image) density estimation, and image
classification. For a full list of hyper-parameters, model
specifications, and example PyTorch \cite{paszke2017automatic} code see the Appendix.

\subsection{Neural Sequence Sorting}

We begin by exploring the problem of neural sequence sorting in order to demonstrate the effectiveness
of our solution. While neural networks outperform humans in many
object detection tasks \cite{he2016deep,szegedy2016inception}, they
generally perform poorly on combinatorial problems such as sorting
\cite{vinyals2015order, vinyals2015pointer} and visual relational learning
\cite{ramapuramnew}. This limitation is due to the exponential growth of potential
solutions in relation to the dimensionality of the input space \cite{tenenbaum2018building}.

$N$ input sequences of length $T$ are generated by sampling a uniform distribution, $\{\bm{X}\}_{i=1}^N =
\{x_1, \ldots, x_t, \ldots, x_T\}_{i=1}^N,\ x_t \sim
U(0, 1)$. The objective of the model,
% : x_{t=1}^T \mapsto y_{t=1}^T
$f_{\theta}$, is to predict a categorical
% output, $\{\bm{Y}\}_{i=1}^N = \{\bm{y}_1, \ldots, \bm{y}_t,
% \ldots, \bm{y}_T \}_{i=1}^N, \bm{y}_t \in \mathbb{R}^{T \times T}$, corresponding to the index
% ,\ \bm{y}_t | \bm{X}_i \sim Cat(\theta), \bm{y}_t \in
% \mathbb{R}^{|\bm{X}_i|},
output, $\{\bm{Y}\}_{i=1}^N, \bm{Y}_i \in \mathbb{R}^{T \times T}$, corresponding to the index
of the sorted input sequence, $\bm{Y}_i =
\text{\emph{sort}}(\bm{X}_i)$. We
provide a single-sample example for $T = 3$ below:

\begin{align}
  \underbrace{[0.6, 0.234, 0.9812]}_{\bm{X}_i} \mapsto \underbrace{[[0, 1, 0], [1, 0, 0], [0, 0, 1]]}_{\bm{Y}_i} \nonumber
\end{align}

We follow \cite{vinyals2015order} and evalute the
all-or-none (called out-of-sequence in \cite{vinyals2015order})
accuracy for all presented models. This metric penalizes an output,
$f_{\theta}(\bm{X}_i)$,  for not predicting the entire sequence in correct order (no
partial-credit),  $\frac{1}{N} \sum_{i=1}^N (f_{\theta}(\bm{X}_i) ==
\bm{Y}_i)$. The reasoning being that a partial sort prediction is not
useful in many cases. Note that larger all-or-none accuracy
implies larger accuracy.

We develop two novel models to address the sorting problem: a simple
feed-forward neural network (Figure \ref{model}-\emph{left}) and a sequential
RNN model (Figure \ref{model}-\emph{right}). The central difference between a
traditional model and the ones in Figure \ref{model}, is the incorporation of a
non-differentiable (hard) function shown in red in both model
diagrams. The dense model % (Figure \ref{model}-\emph{left})
differs
from the RNN model % (Figure \ref{model}-\emph{right})
in that it
receives the entire sample, $x_{t=1}^T$, simultaneously. In contrast, the RNN
processes each value, $x_t$, one at the time, only making a prediction
after the final value, $x_{t=T}$ is received.

% In order to backpropate gradients through these functions we
% use an auxillary DAB network.
During the forward functional evaluations of the model, we directly use the (hard) non-differentiable function's
output for the subsequent layers. The DAB network receives the
same input as the non-differentiable function and caches its output. This cached output is used in the added regularizer
presented in Section \ref{model_sec} in order to allow the DAB to
approximate the non-differentiable function ($\mathcal{L}_{DAB}$ in
Figure \ref{model}). During the backward pass
(dashed lines), the gradients are routed through the DAB instead
of the non-differentiable function. While it is possible to utilize any
non-differentiable function, in this experiment we use the
$\epsilon$-margin signum function from Equation \ref{signum_fn}.
% \vskip -0.1in

% \vskip -0.1in
We contrast our models with state of the art for neural sequence sorting
\cite{vinyals2015pointer, vinyals2015order} \footnote{We report the
  best achieved results (taking pondering (Section \ref{ponder_sec})
  into account) directly from \cite{vinyals2015pointer,
    vinyals2015order}.} and a baseline ELU-Dense
multilayer neural network and demonstrate (Table
\ref{sort_results}) that our model outperforms all
baselines (in some cases by over \textbf{75\%}). Since the \emph{only}
difference between ELU-Dense and Signum-Dense is the
choice of activation, the gains can be attributed to the choice of non-differentiable function that we use in our
model. We believe that the logic of sequence sorting can be simplified using
a function that directly allows binning of intermediary model outputs into
$\{-1, 0, 1\}$, which in turn simplifies implementing a swap
operation in a similar manner as classical Sorting Networks
\cite{batcher1968sorting}.

After observing these significant improvements
over the state of the art in neural sorting \cite{vinyals2015pointer, vinyals2015order}, we attempted to use a soft version of the
$\epsilon$-margin signum function in Equation \ref{signum_fn} (Tanh). We
observed that it performed better than \cite{vinyals2015pointer,
  vinyals2015order} on the 5 and 10 sort problems, but failed to
generalize on the 15-sort problem. The Tanh model resulted in average all-or-none accuracies of
99.3\%, 88.3\% and 1.9\% for the corresponding 5, 10 and 15 sort
problems. The reason for this reduction in performance can be
attributed to the simplification of the problem through
the use of the hard function; eg: instead of learning that all
continuous values in the range of [0.5, 1.0] indicate a swap is needed, the
network only needs to learn that values that are exactly 1.0 indicate
a swap. This reasoning has been extensively applied in vector
quantization (VQ) models such as \cite{gray1984vector,brodsky2000model,van2017neural}.

\vspace{-0.13in}
\subsubsection{Effect of Pondering}\label{ponder_sec}
% \vspace{-0.1in}
% \begin{figure*}
%   \begin{minipage}{0.5\linewidth}
%     \includegraphics[width=\linewidth]{./imgs/5_sort_ponder_acc.png}
%   %   \includegraphics[width=\linewidth]{./imgs/ponder.png}
%   %   \caption{Effect of increasing ponder duration for 5-sort problem. Each
%   % ponder loop was run 5 times.}\label{ponder}
%   \end{minipage}%
%   \begin{minipage}{0.5\linewidth}
%     \includegraphics[width=\linewidth]{./imgs/10_sort_ponder_acc.png}
%   \end{minipage}
%   \caption{Effect of increasing ponder steps for 5 (\emph{left}) and
%     10 (\emph{right}) sort problems. The mean and standard deviation of the maximum test all-or-none accuracy are
%       reported over 5 runs per ponder length.}\label{ponder}
% \end{figure*}%

The model presented in \cite{vinyals2015order} evaluates the effect of
pondering in which they iterate an LSTM with no further inputs. This pondering allows
the model to learn to sort its internal representation. Traditional
sorting algorithms run $O(\log T)$ operations on the $T$ dimensional input
sequence. Iterating the LSTM attempts to parallel this.
We introduce a similar pondering loop into our model and show the
performance benefit in Figure \ref{ponder}; we observe a similar
performance gain, but notice that the benefits decrease after five
pondering iterations.

\begin{figure*}
\begin{minipage}{0.4\textwidth}
    \includegraphics[width=\linewidth]{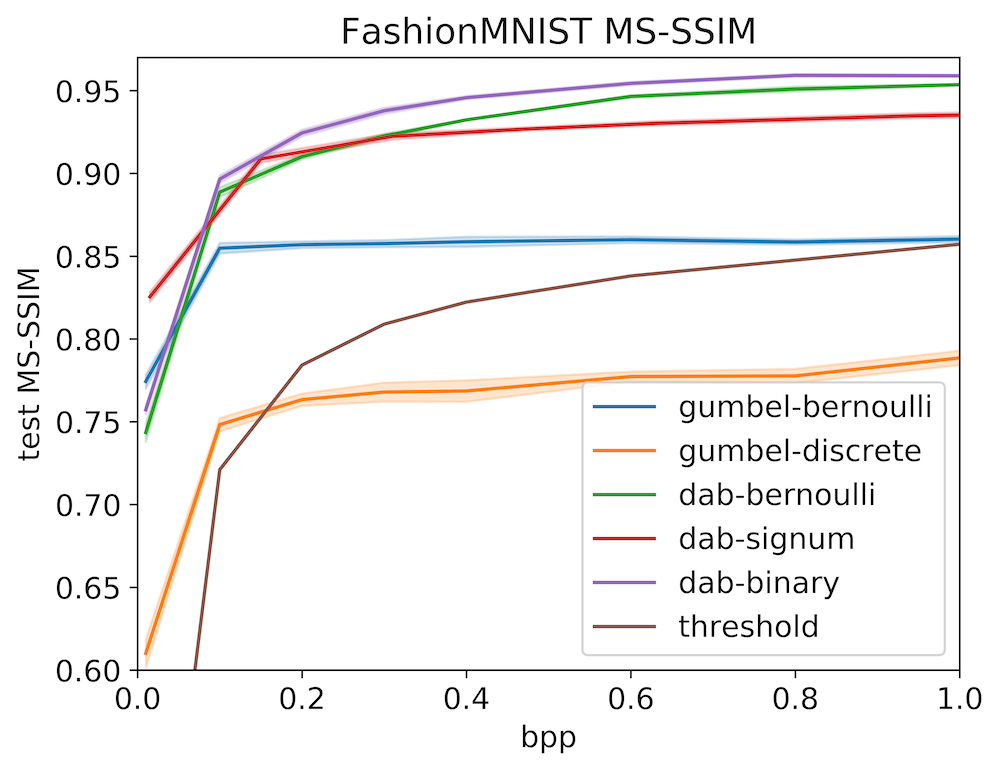}
\end{minipage}%
\begin{minipage}{0.4\textwidth}
  \includegraphics[width=\linewidth]{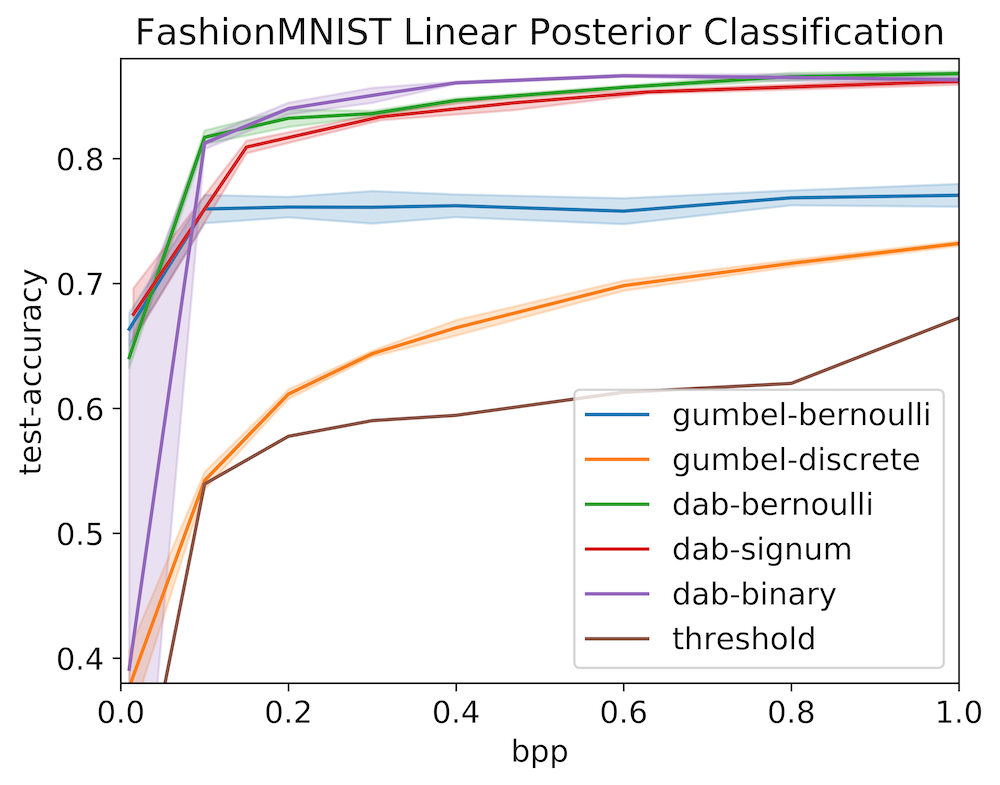}
\end{minipage}%
\begin{minipage}{0.2\textwidth}
  % \vspace{-0.2in}
  \includegraphics[width=\linewidth]{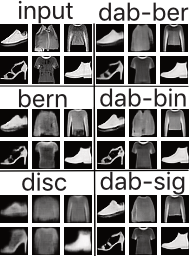}
\end{minipage}%
\\
\begin{minipage}{0.4\textwidth}
  \includegraphics[width=\linewidth]{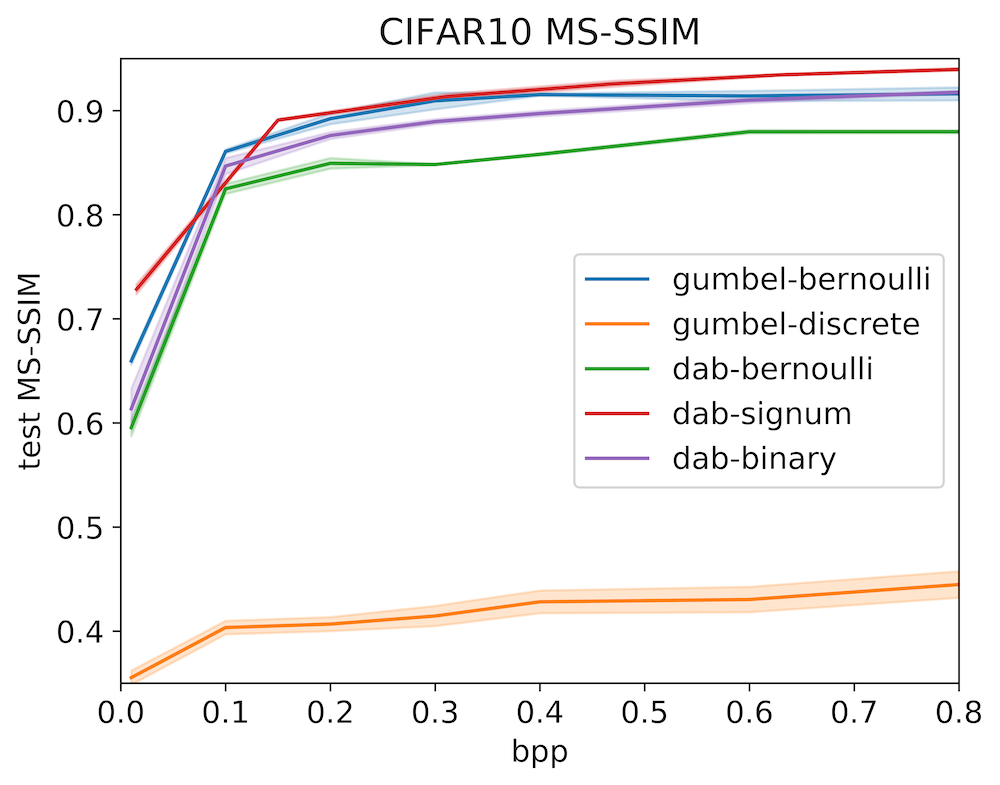}
\end{minipage}%
\begin{minipage}{0.4\textwidth}
  \includegraphics[width=\linewidth]{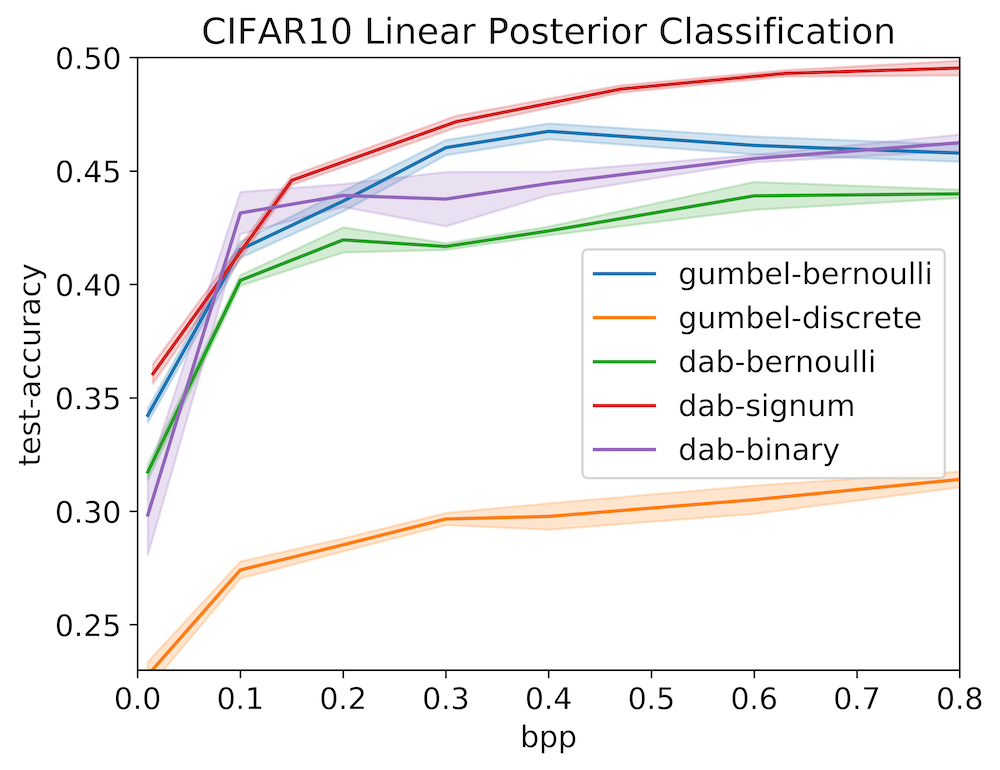}
\end{minipage}%
\begin{minipage}{0.2\textwidth}
  % \vspace{-0.2in}
  \includegraphics[width=\linewidth]{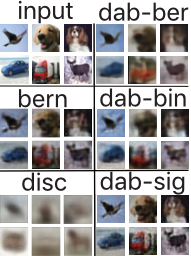}
\end{minipage}%
% \caption{We sweep a range of BPP values for FashionMNIST and CIFAR10,
% performing 5 experiments at each BPP level per model type. Imagenet
% only utilizes BPP=0.00097 due to computational
% restrictions. \emph{Left:} Test MS-SSIM. \emph{Middle:} \textbf{purely
% unsupervised} linear posterior test-classification accuracy; Imagenet
% is compressed to \textbf{786} bits (\textbf{496} for
% \emph{dab-signum}) and yeilds a \textbf{40x} improvement over random
% guessing. \emph{Right:} Test input images and their
% reconstructions.}\label{unsupervised_results}
\caption{We sweep a range of bits-per-pixel (BPP) values for FashionMNIST and CIFAR10,
performing 5 experiments at each BPP level per model
type (results reported as mean $\pm$ std). \emph{Left:} Test Multi-Scale Structural Similarity (MS-SSIM) \cite{wang2003multiscale}. \emph{Middle:} \textbf{Purely
unsupervised} linear posterior test-classification
accuracy. \emph{Right:} Test input images and their
reconstructions at BPP=0.1.}\label{unsupervised_results}
\vskip -0.2in
\end{figure*}

% \begin{figure*}
%   % \vspace{-0.3in}
% \begin{minipage}{0.5\textwidth}
%     \includegraphics[width=\linewidth]{./imgs/imagenet_full_768_msssim.png}
% \end{minipage}%
% \begin{minipage}{0.5\textwidth}
%   \includegraphics[width=\linewidth]{./imgs/imagenet_full_768_linear_sep.png}
% \end{minipage}%
% \caption{Five trials (each) of ImageNet using only BPP=0.00097 due to computational
% restrictions (results reported as mean $\pm$ std). \emph{Left:} Test MS-SSIM \cite{wang2003multiscale}. \emph{Right:} \textbf{Purely
% unsupervised} linear posterior test-classification accuracy; Images
% are compressed from $\mathbb{R}^{512 \times 512}$ to \textbf{786} bits (\textbf{496} for
% \emph{dab-signum} since $786 \approx 496 \log_2(3)$) and yield a \textbf{40x} improvement over random
% guessing (0.001).}\label{imagenet_results}
% \end{figure*}

% \vspace{-0.2in}
\subsection{Unsupervised Discrete Representations}\label{unsup}
% \vspace{-0.2in}

In this experiment, we study the usefulness of learnt unsupervised
representations by latent variable models such as the
Variational Autoencoder (VAE) \cite{kingma2014}. Variational
Autoencoders, coupled with Gumbel relaxed reparameterization methods
\cite{maddison2016concrete,jang2016categorical} enable learning of
compact binary latent representations. Given an input random variable
$x \sim p(x)$, VAEs posit an approximate
posterior, $q_{\phi}(z_2 | x )$, over a latent variable,
$z_2$, and maximize the Evidence Lower BOund (ELBO). We contrast the VAE
ELBO with our optimization objective below \footnote{Note that the
  backpropagation step for the DAB follows the same logic as presented
  earlier in Section \ref{model_sec}.}:

\vspace{-0.1in}
{\centering
  % \begin{table}[H]
%     \begin{center}
% {\renewcommand{\arraystretch}{1.3}%
% \begin{tabular}{|l|l|}
% \hline
% \multicolumn{1}{|c|}{\textbf{VAE}} & \multicolumn{1}{c|}{\textbf{DAB}} \\ \hline
% {$\!\begin{aligned}
%   \mathbb{E}_{q} [\log p_{\theta} (x | z) ] -
%   D_{KL}[q_{\phi}(z| x) || p(z)]
% \end{aligned}$}             & {$\!\begin{aligned}
%   \mathbb{E}_{q} [\log p_{\theta} (x | z) ] +
%   \gamma \log q(z_{\text{hard}} | f_{\phi}(z^{i-1}))
% \end{aligned}$}             \\ \hline
% \end{tabular}
% }
% \end{center}
% \end{table}
  \begin{table}[H]
    \begin{center}
{\renewcommand{\arraystretch}{1.3}%
\begin{tabular}{|l|}
\hline
\textbf{VAE}: {$\!\begin{aligned}
   \mathbb{E}_{q} [\log p_{\theta} (x | z_2) ] -
   D_{KL}[q_{\phi}(z_2| x) || p(z)]
 \end{aligned}$}  \\ \hline
\textbf{DAB}: {$\!\begin{aligned}
   \mathbb{E}_{q} [\log p_{\theta} (x | z_{\text{hard}}) ] +
   \mathbb{E}_{q} [\gamma \log q(z_{\text{hard}} | l_{\phi}(z_1))]
 \end{aligned}$} \\ \hline
\end{tabular}
}
\end{center}
\end{table}
}
\vspace{-0.25in}

In the case of the DAB model, $q(z_{\text{hard}} | l_{\phi}(z_1)) \sim
\mathcal{N}(l_{\phi}(z_1) , I)$. This formulation is equivalent to the
one from Section \ref{model_sec} since the log-likelihood of a
Gaussian distribution (evaluated on a sample $z_{\text{hard}}$) is proportional to the L2-loss (Appendix Section
\ref{regularizer}). The specific functional value of $z_{\text{hard}}$ is based on
the type of non-differential function used and is listed in Table \ref{variants}.

Good latent representions should not only be compact
(in terms of bits-per-pixel), but also useful as a mechanism to
produce a more separable representation space\footnote{This differs
  from recent work on disentangled representations \cite{locatello2018challenging} which
  necessitate each dimension of the latent variable independently
  control a factor of variation in the original data space.}, $z_2$. In addition, given a
latent representation the model should be able to reconstruct the
original sample well. We demonstrate the usefulness of non-differentiable
functions to both these objectives through the use of two
metrics: the MS-SSIM \cite{wang2003multiscale} and linear
classification of posterior samples. The MS-SSIM is a metric % typically used in
% compression related studies and
that provides a sense of how
similar (in structure) the reconstructed image is to the
original. Linear classification of posterior samples provides us with
an evaluation of separable latent representations: a useful property
for models that use the latent representation in downstream tasks. Importantly, we \textbf{do not}
specifically train the model to induce better linearly separability as
that would necessitate the use of supervision.
% We posit that both are
% necessary in an unsupervised learning scenario and present both
% metrics in Figure \ref{unsupervised_results}.

% \begin{figure}[H]
% \begin{minipage}{0.5\textwidth}
%     \includegraphics[width=\linewidth]{./imgs/msssim_ts_v1_cifar10.png}
% \end{minipage}%
% \begin{minipage}{0.5\textwidth}
%   \includegraphics[width=\linewidth]{./imgs/linear_test_acc_ts_v1_cifar10.png}
% \end{minipage}%
% \caption{\emph{Left:} Test MS-SSIM for CIFAR10. \emph{Right:} Test linear posterior classification accuracy for CIFAR10}\label{unsupervised_results_cifar}
% \end{figure}

% \begin{figure}[H]
% \begin{minipage}{0.5\textwidth}
%     \includegraphics[width=\linewidth]{./imgs/imagenet_768_msssim.png}
% \end{minipage}%
% \begin{minipage}{0.5\textwidth}
%   \includegraphics[width=\linewidth]{./imgs/imagenet_768_msssim.png}
% \end{minipage}%
% \caption{\emph{Left:} Test MS-SSIM for Imagenet at BPP=0.00097. \emph{Right:}
%   Test linear posterior classification accuracy for Imagenet at BPP=0.00097}\label{unsupervised_results_imagenet}
% \end{figure}

In Figure \ref{unsupervised_results} and Table \ref{imagenet_results}, we contrast our models (\emph{dab-*})
against state of the art \emph{bernoulli} and \emph{discrete}
% gumbel-reparameterized\\ \\ \\ \\ \\ \\ \\models
% \ \\ \\ \\ \\ \\ \\ \\
relaxed gumbel-reparameterized models
\cite{maddison2016concrete,jang2016categorical} and a naive
downsample, binary-threshold and classify using optimal
threshold solution (\emph{threshold}). The variants are summarized in Table \ref{variants}
below: % $we utilize below:

\vskip -0.1in
 {\centering
  \begin{table}[H]
    \begin{center}
\begin{tabular}{l|l|}
\cline{2-2}
                                             & \multicolumn{1}{c|}{\textbf{Functional Form}} \\ \hline
\multicolumn{1}{|l|}{\textbf{dab-bernoulli}} & \begin{tabular}[c]{@{}l@{}}Sample from
                                               non-reparameterized
                                               \\ distribution: $z_{\text{hard}} \sim Bern(l_{\theta_1}(x))$.\end{tabular}                     \\ \hline
\multicolumn{1}{|l|}{\textbf{dab-binary}}    & {$\!\begin{aligned}
  bin(z_1) &= \begin{cases}
      1 & z_1 \geq \text{mean}(z_1) \\
      0 & z_1 < \text{mean}(z_1)
   \end{cases}
\end{aligned}$}                      \\ \hline
\multicolumn{1}{|l|}{\textbf{dab-signum}}    & \begin{tabular}[c]{@{}l@{}}Equation
                                               \ref{signum_fn}.
                                               BPP is scaled by $\log_2(3)$\\ due to trinary
                                               representation.\end{tabular}
  \\ \hline
\multicolumn{1}{|l|}{\textbf{threshold}} &  \begin{tabular}[c]{@{}l@{}}bilinear(x, BPP),
                                           threshold(x, $\tau$) and\\ linearly classify for the
                                           best $\tau$.\end{tabular}
                                            \\ \hline
\end{tabular}
\end{center}
\caption{Variants of activations used in experiment.}\label{variants}
\end{table}
}
\vspace{-0.15in}
% {\centering
%   \begin{table}[H]
%     \begin{center}
% \begin{tabular}{l|l|}
% \cline{2-2}
%                                              & \multicolumn{1}{c|}{\textbf{Functional Form}} \\ \hline
% \multicolumn{1}{|l|}{\textbf{dab-bernoulli}} & Sample from
%                                                non-reparameterized
%                                                distribution: $z \sim Bern(f_{\theta}^1(x^*))$.                     \\ \hline
% \multicolumn{1}{|l|}{\textbf{dab-binary}}    & {$\!\begin{aligned}
%   bin(x) &= \begin{cases}
%       1 & x \geq \text{mean}(x) \\
%       0 & x < \text{mean}(x)
%    \end{cases}
% \end{aligned}$}                      \\ \hline
% \multicolumn{1}{|l|}{\textbf{dab-signum}}    & Equation
%                                                \ref{signum_fn}.
%                                                BPP is scaled by $\log_2(3)$
%                                                due to trinary
%                                                representation.
%   \\ \hline
% \multicolumn{1}{|l|}{\textbf{threshold}} &  bilinear(x, BPP),
%                                            threshold(x, $\tau$) and
%                                            linearly classify for the
%                                            best $\tau$. \\ \hline
% \end{tabular}
% \end{center}
% \end{table}
% }

% \vspace{-0.1in}
% The
% \emph{proxy-bernouilli} model simply utilizes a non-differentiable
% version of the reparameterization, i.e. $z \sim
% q_{\phi}(z | x)$ instead of its reparameterized variant.

% \scalebox{0.75}{\parbox{1.0\linewidth}{%

We begin by using the training set of Fashion MNIST% \cite{xiao2017/online}
, CIFAR10, and ImageNet to train the baseline bernoulli and discrete
VAEs as well as the models with the non-differentiable functions (\emph{dab-*}) presented above. We train \emph{five models
per level of bpp} for FashionMNIST and CIFAR10 and evaluate the MS-SSIM and linear classification
accuracy at each point. We repeat the same, but only for bpp=0.00097
for Imagenet ($512 \times 512 \times 3$) due to computational
restrictions. Each epoch of training at this resolution takes
approximately 1.5 hours on 8 V-100 GPUs. Note that
  bpp=0.00097 requires a matrix that projects into a 786 dimensional
  space. Increasing this dimension substantially increases the
  parameters of the network. The linear classifier is trained on the same
training dataset\footnote{We use the encoded latent
  representation as input to the linear classifier.} after the
completion of training of the generative model. We present the mean and standard deviation
results in Figure \ref{unsupervised_results} and Table
\ref{imagenet_results} for all three datasets.

% \vspace{-0.1in}
\begin{table}[H]
  {\renewcommand{\arraystretch}{1.3}%
    \begin{center}
  \begin{adjustbox}{width=\columnwidth}
\begin{tabular}{|l|l|l|}
\hline
\textbf{\begin{tabular}[c]{@{}l@{}}Imagenet BPP = 0.00097\\ (768 dimensional latent)\end{tabular}} & \textbf{MS-SSIM}            & \textbf{Linear Separability} \\ \hline
Gumbel-Bernoulli                                                                                   & 0.295 +/- 0.00058           & 0.0405 +/- 0.00035           \\ \hline
DAB-Signum $\dagger$                                                                                         & \textbf{0.296 +/- 0.00063} & \textbf{0.0430 +/- 0.00068}  \\ \hline
DAB-Bernoulli                                                                                      & 0.293 +/- 0.00051           & 0.0387 +/- 0.00022           \\ \hline
DAB-Binary                                                                                         & 0.292 +/- 0.00062           & 0.0356 +/- 0.00092           \\ \hline
\end{tabular}
  \end{adjustbox}
\end{center}
}
\caption{Five trials (each) of ImageNet using only BPP=0.00097
  (results reported as mean $\pm$ std). Images are compressed from
  $\mathbb{R}^{512 \times 512 \times 3}$ to \textbf{786} bits ($\dagger$ \textbf{496} for
  \emph{dab-signum} since $786 \approx 496 \log_23$) and yield a
  \textbf{43x} improvement over random guessing (0.001). Full test
  curves in Appendix Section \ref{imagenet_curves}.}\label{imagenet_results}
\end{table}
% }}
\vspace{-0.1in}
We observe that our
models perform better in terms of test-reconstruction (MS-SSIM) and also
provide a more separable latent representation (in terms of linear test
accuracy). We observe either \emph{dab-signum} or \emph{dab-binary} performing better than all variants across all datasets. Since
\emph{only} the activation is being changed, the benefit can be directly attributed
to the use of the non-differentiable functions used as activations.
Since the DAB-decoder only operates over discrete inputs,
it drastically simplifies the learning problem for this network. This contrasts the
Gumbel-Softmax estimator which slowly anneals a continuous
distribution to a discrete one over the training process. This
validates the core tenant of DAB: use discrete (non-differentiable)
outputs during the forward functional evaluations, but provide a smooth
K-Lipschitz gradient during backpropagation.

\vspace{-0.2in}
\subsubsection{Contrasting State of the Art Discrete Estimators}\label{other_estimators_exp}
% While REBAR \cite{tucker2017rebar} and
% RELAX \cite{grathwohl2017backpropagation} are unbiased
% estimators, DAB makes no such guarantees. However, in practice we
% observe that DAB training is simpler,
% having one unique hyper-parameter (the regularizer weighting, $\gamma$,
% in Equation \ref{dab_vae}) in contrast to RELAX
% which has three hyper-parameters as well as two optimizers with different
% parameters. % DAB converges faster and provides
% % better empirical performance as we will soon demonstrate.

In order to relate this work to newer \emph{state of the art} discrete
estimators \cite{tucker2017rebar, grathwohl2017backpropagation}, we
parallel the simple VAE experiments used for RELAX
\cite{grathwohl2017backpropagation} and REBAR \cite{tucker2017rebar}.
The experiment proposed in \cite{tucker2017rebar,
  grathwohl2017backpropagation} is to estimate a (variational) density
model for Binarized MNIST and
Binarized Omniglot using a latent variable posterior distribution of 200 Bernouilli
random variables. This task is challenging for neural
networks that learn with gradient descent, as quantization (of the
forward functional evaluations) removes the subtle directional information from the gradients. Thus, all of the proposed
relaxed estimators \cite{maddison2016concrete,jang2016categorical,tucker2017rebar,grathwohl2017backpropagation}
use some form of annealed continuous distribution (such as the Gumbel
distribution) during their forward functional evaluations.  These distributions provide the model with
continuous gradient information to update parameters during
backpropagation. Over time, these continuous distributions are annealed towards the
desired discrete representation, albeit sometimes with large variance
\cite{tucker2017rebar}. In contrast, DABs always use discrete
outputs during forward functional evaluations, while providing a
smooth, K-Lipschitz gradient to enable learning. This
approach allows the decoder in the VAE model to restrict itself to the range of
discrete numerical values that the problem specifies.

% We use Binarized
% MNIST [cite laroche] and Omniglot and list the best
% achieved negative evidence lower bound (ELBO) for all models in
% Tables \ref{binarized_mnist} and \ref{binarized_omniglot} (larger
% values of the variational bound being better).
As in \cite{tucker2017rebar,grathwohl2017backpropagation}, we use a
single hidden layer model with ReLU activations and 200 latent Bernouilli random
variables. Adam is used as an optimizer with a learning rate of
3e-4 and $\gamma$ from Equation \ref{dab_loss} is fixed to 10. In contrast to Experiment
\ref{unsup}, we optimize the following objective:

\scalebox{0.95}{\parbox{1.0\linewidth}{%
\begin{align}
  \begin{split}
\hspace{-0.2in}\textbf{DAB-VAE}:
   \mathbb{E}_{q} [\log p_{\theta} (x | z_{\text{hard}}) ] -
   D_{KL}[q_{\phi}(l_{\phi}(z_1)| x) || p(z)] \\+
   \mathbb{E}_{q}[\gamma \log q(z_{\text{hard}} |
   l_{\phi}(z_1))] \label{dab_vae}
   \end{split}
\end{align}
}}

\noindent The objective in Equation \ref{dab_vae} ensures that the the DAB-VAE
compress the latent variable in a manner similar to REBAR and RELAX.

\begin{table*}
  {\renewcommand{\arraystretch}{1.0}%
\begin{tabularx}{\textwidth}{X|l|l|X|}
\cline{2-4}
\makecell{CIFAR10 \\Test-Accuracy}                & \textbf{Mean}  & \textbf{+/- Std} & \textbf{Functional Form} \\ \hline
\multicolumn{1}{|l|}{\textbf{Baseline}}       & 92.87\% & 0.06\%        & Identity($z_1$)                \\ \hline
\multicolumn{1}{|l|}{\textbf{Signum}}         & 91.95\% & 0.07\%       & Equation \ref{signum_fn}                \\ \hline
\multicolumn{1}{|l|}{\textbf{Sort}}     & \textbf{92.93\%} & \textbf{0.1\%}       &
                                                                  sort-row($z_1$)
                                                                  $\oplus$
                                                                  sort-col($z_1$)\\
  \hline
\multicolumn{1}{|l|}{\textbf{Topk}}           & 92.21\% & 0.14\%
                                                                                      &
                                                                                        (sort-row($z_1$) $\oplus$ sort-col($z_1$))[0:k]                \\ \hline
\multicolumn{1}{|l|}{\textbf{K-Means}} & 91.97\% & 0.16\%       & kmeans($z_1$, k=10) \\ \hline
\end{tabularx}
}
\caption{CIFAR10 test-accuracy over five trials for each
  row. $\oplus$ is a concatenation.}\label{cifar10_acc}
\end{table*}

{\centering
  % \scalebox{0.85}{\parbox{1.0\linewidth}{%
    \begin{table}[H]
      {\renewcommand{\arraystretch}{1.3}%
        \begin{center}
          \begin{adjustbox}{width=\columnwidth}
        \begin{tabular}{c|c|c|c|c|}
\cline{2-5}
\multicolumn{1}{l|}{}            & \begin{tabular}[c]{@{}c@{}}Binarized \\ MNIST\end{tabular} & \begin{tabular}[c]{@{}c@{}}Binarized \\ Omniglot\end{tabular} & \begin{tabular}[c]{@{}c@{}}Epochs \\ Binarized\\ MNIST\end{tabular} & \begin{tabular}[c]{@{}c@{}}Epochs\\ Binarized\\ Omniglot\end{tabular} \\ \hline
\multicolumn{1}{|c|}{REBAR \cite{tucker2017rebar}}      & -111.12                                           & -127.51                                              & \textbf{331}                                                                 & \textbf{368}                                                                   \\ \hline
\multicolumn{1}{|c|}{RELAX \cite{grathwohl2017backpropagation}}      & -119.19                                                    & -128.20                                                       & \emph{-}\emph{-}\emph{-}                                                                   & \emph{-}\emph{-}\emph{-}                                                                     \\ \hline
\multicolumn{1}{|c|}{DAB (ours)} & \textbf{-109.59}                                                    & \textbf{-125.19}                                                        & 9933                                                                 & 2366                                                                   \\ \hline
        \end{tabular}
        \end{adjustbox}
        \end{center}
}
\caption{Binarized MNIST \& Omniglot test variational lower bound
  (ELBO) in nats and training epochs to
  convergence. \emph{-}\emph{-}\emph{-} indicates non-reported
  values.}\label{binarized_results}
\end{table}
% }}
}

% We perform five trials per model type and report the mean and standard
% deviation of the test variational lower bound ( Equation \ref{dab_vae}
% without $\mathbb{E}_{q}[\gamma \log q(z_{\text{hard}} |
% f_{\phi}(z^{i-1}))]$ ) for Binarized MNIST and
% Omniglot in Table \ref{binarized_results} (larger values being better). In
% addition, we observe that DAB converges in $\approx$ \textbf{200}
% epochs for Binarized MNIST vs. 531 epochs for RELAX and 857 for
% REBAR.
We report the best test variational lower bound (Equation \ref{dab_vae}
without $\mathbb{E}_{q}[\gamma \log q(z_{\text{hard}} |
l_{\phi}(z_1))]$ to provide a meaningful comparison) for Binarized MNIST and
Omniglot in Table \ref{binarized_results} (larger ELBO values being
better) \footnote{Full test curves in Appendix
  Section \ref{contrasting_relax_rebar}.}. The same table also
provides the number of training epochs needed for each model to
converge to their reported best value. While DAB
takes longer to converge, both REBAR and
RELAX begin to overfit and continued training does not improve the
bound. We observed that the generalization gap for DAB (1-2 nats) was
smaller than RELAX and REBAR (5-10 nats). This improvement is because the
decoder, $\log p_{\theta}(x|z_{\text{hard}})$, is required to reconstruct (at
training and testing) the input
sample, $x$, using a latent variable sampled from a stochastic,
non-differentiable function, $z_{\text{hard}} \sim Bern(l_{\theta_1}(x))$. DAB
outperforms both REBAR and RELAX on both the Binarized Omniglot and
Binarized MNIST problems; the relative simplicity of training makes it a strong candidate for
applications requiring non-differentiable functions. Adding a
learning rate scheduler to the DAB based training is likely to improve
convergence time, however this is left to future work.

\subsection{Image Classification}
\begin{figure*}
\begin{minipage}{0.33\textwidth}
  \includegraphics[width=\linewidth]{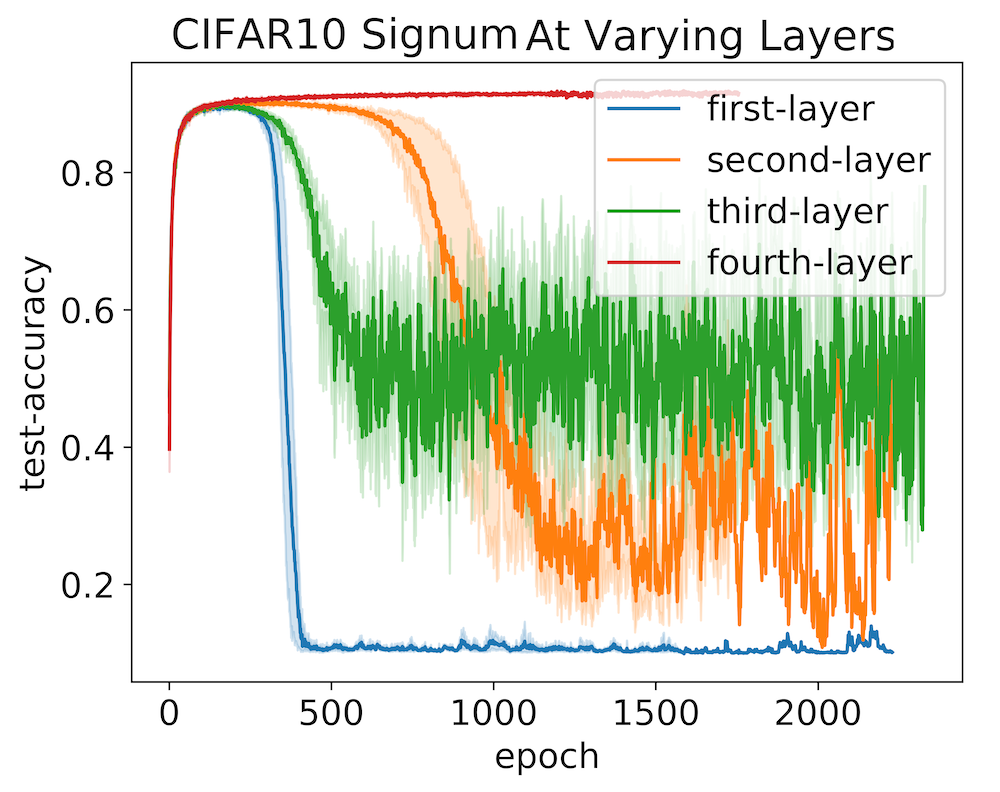}
\end{minipage}%
\begin{minipage}{0.33\textwidth}
  \includegraphics[width=\linewidth]{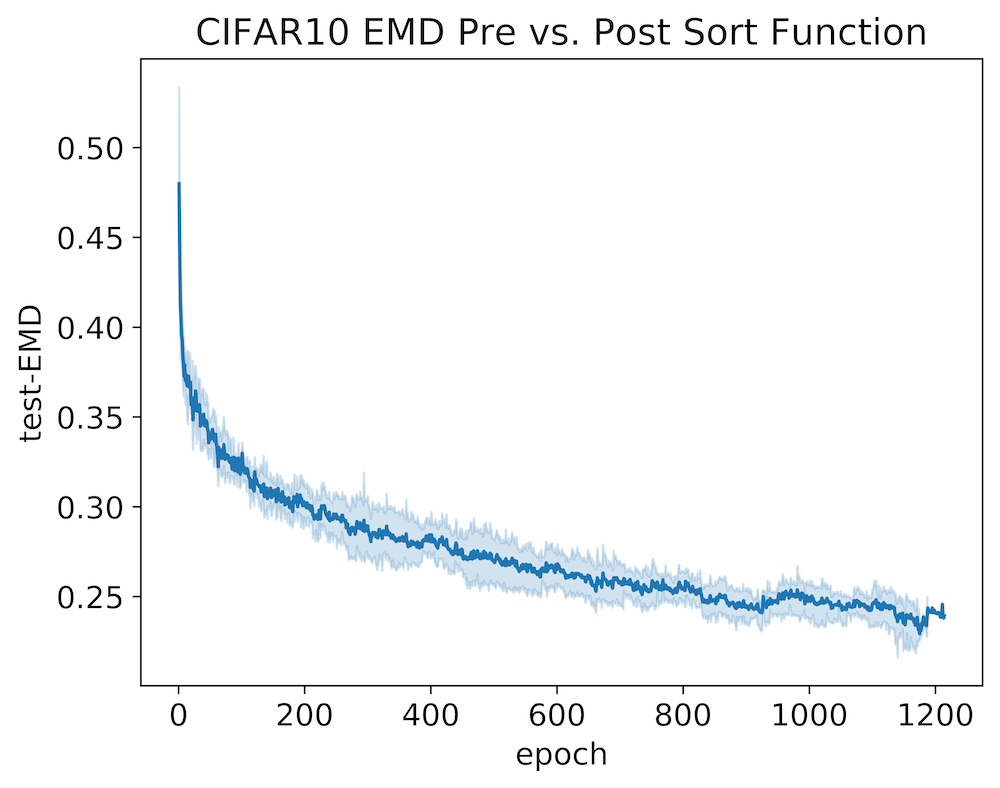}
\end{minipage}%
\begin{minipage}{0.33\textwidth}
  \includegraphics[width=\linewidth]{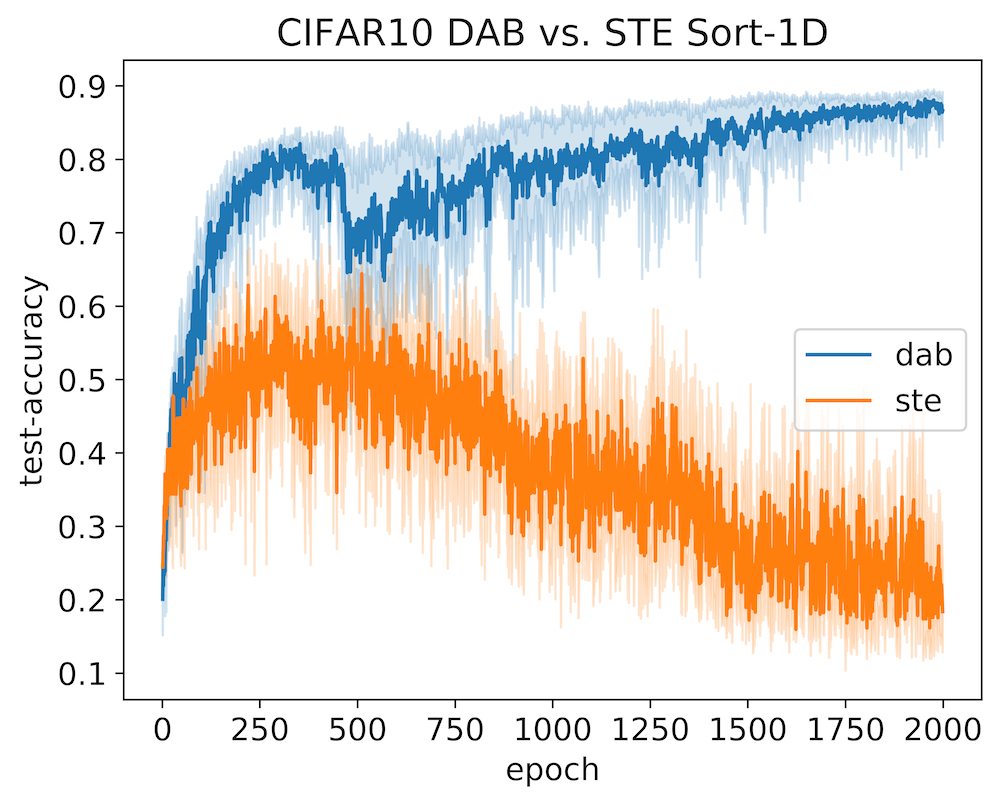}
\end{minipage}%
\caption{\emph{Left:} Signum non-differentiable function evaluated at
  different sections of a Resnet18 model. \emph{Middle:} Earth
  mover distance between input to non-differentiable function
  and output of non-differentiable function. \emph{Right:} CIFAR10
  test accuracy for DAB vs. Straight-Through-Estimator using
  Sort-1D.}\label{cifar10_ablation}
% \vspace{-0.15in}
\end{figure*}

This experiment evaluates how well a DAB enhanced model performs in
classifying images of CIFAR10 using a Resnet18 model tailored to
operate on $\mathbb{R}^{32 \times 32 \times 3}$ images. We evaluate a
variety of non-differentiable functions and present their test
accuracy and standard deviation in Table \ref{cifar10_acc}. We observe
that utilizing a sort as the final activation in the Resnet18 model
improves upon the vanilla model (\emph{Baseline}) by 0.1\%.
While these results show that DAB outperforms the baseline Resnet18 model, the difference is small. In
contrast, when we used the same non-differentiable function in a
simpler multi-layer dense model for the same problem, we observed a larger difference (\textbf{$\approx$10\%}) between the
test-accuracies. We attribute this to the quantization based regularization effect
induced by the choice of non-differentiable activation.

Many of the tested non-differentiable activations in Table
\ref{cifar10_acc} perform equivalently to the state of the art
(\emph{Baseline}). DAB enables the exploration of novel networks with
unconvential layers, including ones (such as MergeSort) which
generalize to arbitrary numerical values. These unconvential
activations can also be used for auxiliary tasks (eg: \cite{van2017neural}).

% We observe
% that non-differentiable functions do not reduce the penultimate
% accuracy in comparison to the baseline model.
% We observe that most of the
% models performs within tolerance of each other even though the
% provided functions are non-differentiable.
% \vspace{-0.1in}
\subsubsection{Classification Ablation / Case Studies}

\textbf{Layer Placement}: In order to validate where to place the non-differentiable
function within the Resnet18 architecture, we perform an ablation study. Since the Resnet18 model has four
residual blocks, we place the non-differentiable function at the output
of each block and train with each configuration 5 times (Figure
\ref{cifar10_ablation}-\emph{left}). We observe that the network
remains stable throughout training when the
non-differentiable function is after he fourth layer and use this
configuration for all
experiments presented in Table \ref{cifar10_acc}. \\ % This is likely because networks typically learn low level
% Haar like filters at initial layers and enacting a complex, non-differentiable
% function at an initial layer destroys the coherence during the
% learning process. \\

% \begin{figure}[H]
% \begin{minipage}{\linewidth}
%   \includegraphics[width=\linewidth]{./imgs/cifarv4_sort_emd.png}
% \end{minipage}%
% \caption{\emph{Left:} Signum non-differentiable function evaluated at
%   different sections of a Resnet18 model. \emph{Middle:} Earth
%   mover distance between input layer to non-differentiable function
%   and output of non-differentiable function. \emph{Right:} CIFAR10
%   test accuracy for DAB vs. Straight-Through-Estimator using Sort-1D.}\label{cifar10_ablation}
% \end{figure}

% \begin{figure}[H]
% % \begin{minipage}{\linewidth}
% %   \includegraphics[width=\linewidth]{./imgs/cifar10_varying_layers_signum_mixup.png}
% % \end{minipage}%
% \begin{minipage}{\linewidth}
%   \includegraphics[width=\linewidth]{./imgs/cifarv4_sort_emd.png}
% \end{minipage}%
% % \begin{minipage}{0.33\textwidth}
% %   \includegraphics[width=\linewidth]{./imgs/cifarv4_dab_vs_ste.png}
% % \end{minipage}%
% \caption{Earth
%   mover distance between input layer to non-differentiable function
%   and output of non-differentiable function.}\label{cifar10_ablation}
% \end{figure}
% \vspace{-0.05in}
\noindent\textbf{Conditioning of Preceding Layer}: We utilize the \textbf{Sort} non-differentiable function shown in Table
\ref{cifar10_acc} to explore the effect of the regularizer
introduced in Equation \ref{dab_loss}. We calculate the empirical
earth mover distance between the input to the
non-differentiable function ($z_1$ in Figure \ref{graphical_model_v2})
and its output ($z_{\text{hard}}$ in Figure
\ref{graphical_model_v2}). We repeat the experiment five times and report
the mean and standard deviation in Figure
\ref{cifar10_ablation}-\emph{middle}. The regularizer conditions the
input layer  to produce partially sorted values, as demonstrated by the decrease in the test
EMD over time.

% \begin{figure}[H]
% \begin{minipage}{\linewidth}
%   \includegraphics[width=\linewidth]{./imgs/cifarv4_dab_vs_ste.png}
% \end{minipage}%
% \caption{\emph{Left:} Signum non-differentiable function evaluated at
%   different sections of a Resnet18 model. \emph{Middle:} Earth
%   mover distance between input layer to non-differentiable function
%   and output of non-differentiable function. \emph{Right:} CIFAR10
%   test accuracy for DAB vs. Straight-Through-Estimator using Sort-1D.}\label{cifar10_ablation}
% \end{figure}

% \begin{figure}[H]
% % \begin{minipage}{\linewidth}
% %   \includegraphics[width=\linewidth]{./imgs/cifar10_varying_layers_signum_mixup.png}
% % \end{minipage}%
% % \begin{minipage}{0.33\textwidth}
% %   \includegraphics[width=\linewidth]{./imgs/cifarv4_sort_emd.png}
% % \end{minipage}%
% \begin{minipage}{\linewidth}
%   \includegraphics[width=\linewidth]{./imgs/cifarv4_dab_vs_ste.png}
% \end{minipage}%
% \caption{ CIFAR10 test accuracy for DAB vs. Straight-Through-Estimator using Sort-1D.}\label{cifar10_ablation}
% \end{figure}
% \vspace{-0.05in}
\noindent\textbf{Contrasting the STE}: The straight-through-estimator (STE) was originally used to bypass differentiating
through a simple argmax operator \cite{bengio2013estimating}, however,
here we analyze how well it performs when handling a complex operand
such as sorting. Since the STE cannot operate over transformations
that vary in dimensionality, we use a simplified version of the sort
operator from the previous experiment. Instead of sorting the rows and
columns as in Table \ref{cifar10_acc}, we simply flatten the feature
map and run a single sort operation. This allows us to use the STE
in this scenario. We observe in Figure \ref{cifar10_ablation}-\emph{right} that DAB
clearly outperforms the STE.

%%% Local Variables:
%%% mode: latex
%%% TeX-master: "main"
%%% End:

% -------------------------------------------------------------------------
% \vspace{-0.2in}
\section{Discussion}
% \vspace{-0.15in}

Extensive research in machine learning has focused on discovering new
(sub-)differentiable non-linearities to use within neural
networks \cite{hahnloser2000digital,klambauer2017self,
  ramachandran2017searching}. In this work, we demonstrate a novel method to allow for the incorporation of
simple non-differentiable functions within neural networks and empirically
demonstrate their benefit through a variety of experiments using
a handful of non-differentiable operators, such as \emph{kmeans},
\emph{sort}, and \emph{signum}. Rather than manually deriving
sub-differentiable solutions (eg: \cite{grefenstette2015learning}),
using the Straight-Through-Estimator (eg: \cite{van2017neural}) or relying on REINFORCE, we train using a neural network to
learn a smooth approximation to the non-differentiable function. This work opens up
the use of more complex non-differentiable operators
within neural network pipelines.
% and
% utilize its gradients to update previous layers

{\small
\bibliographystyle{ieee}
\bibliography{bibliography}
}

\newpage
\onecolumn
\section{Appendix}

% Default fixed font does not support bold face
\DeclareFixedFont{\ttb}{T1}{txtt}{bx}{n}{12} % for bold
\DeclareFixedFont{\ttm}{T1}{txtt}{m}{n}{12}  % for normal

% Custom colors
% \usepackage{color}
\definecolor{deepblue}{rgb}{0,0,0.5}
\definecolor{deepred}{rgb}{0.6,0,0}
\definecolor{deepgreen}{rgb}{0,0.5,0}

% Python style for highlighting
\newcommand\pythonstyle{\lstset{
language=Python,
% basicstyle=\ttm\tiny,
basicstyle=\ttfamily\tiny,
ndkeywords={BaseHardFn,forward,_hard_fn,backward},
ndkeywordstyle=\tiny\color{deepred},
otherkeywords={None, True, False, self, \@staticmethod},             % Add keywords here
keywordstyle=\tiny\color{deepblue},
emph={MyClass,__init__},          % Custom highlighting
emphstyle=\tiny\color{deepred},    % Custom highlighting style
stringstyle=\color{deepgreen},
commentstyle=\color{deepgreen},
frame=tb,                         % Any extra options here
showstringspaces=false            %
}}

% Python environment
\lstnewenvironment{python}[1][]
{
\pythonstyle
\lstset{#1}
}
{}

% Python for external files
\newcommand\pythonexternal[2][]{{
\pythonstyle
\lstinputlisting[#1]{#2}}}

% Python for inline
\newcommand\pythoninline[1]{{\pythonstyle\lstinline!#1!}}

\subsection{Simple Pytorch Implementation}
We provide an example of the base class for any hard function along
with an example of the $\epsilon$-margin signum operand (Equation \ref{signum_fn}) below. The \emph{BaseHardFn}
accepts the input tensor \emph{x} along with the DAB approximation
(\emph{soft\_y}). Coupling this with the DAB loss (Equation
\ref{regularizer}) provides a basic interface for using DABs with any
model.

% {\centering
  % \includegraphics[width=\linewidth]{./imgs/code.png}
  \begin{python}
    class BaseHardFn(torch.autograd.Function):
    @staticmethod
    def forward(ctx, x, soft_y, hard_fn, *args):
        """ Runs the hard function for forward, cache the output and returns.
            All hard functions should inherit from this, it implements the autograd override.

        :param ctx: pytorch context, automatically passed in.
        :param x: input tensor.
        :param soft_y: forward pass output (logits) of DAB approximator network.
        :param hard_fn: to be passed in from derived class.
        :param args: list of args to pass to hard function.
        :returns: hard_fn(tensor), backward pass using DAB.
        :rtype: torch.Tensor

        """
        hard = hard_fn(x, *args)
        saveable_args = list([a for a in args if isinstance(a, torch.Tensor)])
        ctx.save_for_backward(x, soft_y, *saveable_args)
        return hard

    @staticmethod
    def _hard_fn(x, *args):
        raise NotImplementedError("implement _hard_fn in derived class")

    @staticmethod
    def backward(ctx, grad_out):
        """ Returns DAB derivative.

        :param ctx: pytorch context, automatically passed in.
        :param grad_out: grads coming into layer
        :returns: dab_grad(tensor)
        :rtype: torch.Tensor

        """
        x, soft_y, *args = ctx.saved_tensors
        with torch.enable_grad():
            grad = torch.autograd.grad(outputs=soft_y, inputs=x,
                                       grad_outputs=grad_out,
                                       retain_graph=True)

        return grad[0], None, None, None

class SignumWithMargin(BaseHardFn):
    @staticmethod
    def _hard_fn(x, *args):
        """ x[x < -eps] = -1
            x[x > +eps] = 1
            else x = 0

        :param x: input tensor
        :param args: list of args with 0th element being eps
        :returns: signum(tensor)
        :rtype: torch.Tensor

        """
        eps = args[0] if len(args) > 0 else 0.5
        sig = torch.zeros_like(x)
        sig[x < -eps] = -1
        sig[x > eps] = 1
        return sig

    @staticmethod
    def forward(ctx, x, soft_y, *args):
        return BaseHardFn.forward(ctx, x, soft_y, SignumWithMargin._hard_fn, *args)
  \end{python}
% }

\subsection{Contrasting RELAX and REBAR}\label{contrasting_relax_rebar}
\vspace{-0.2in}

\begin{figure}[H]
  \begin{minipage}{0.5\textwidth}
    \includegraphics[width=\linewidth]{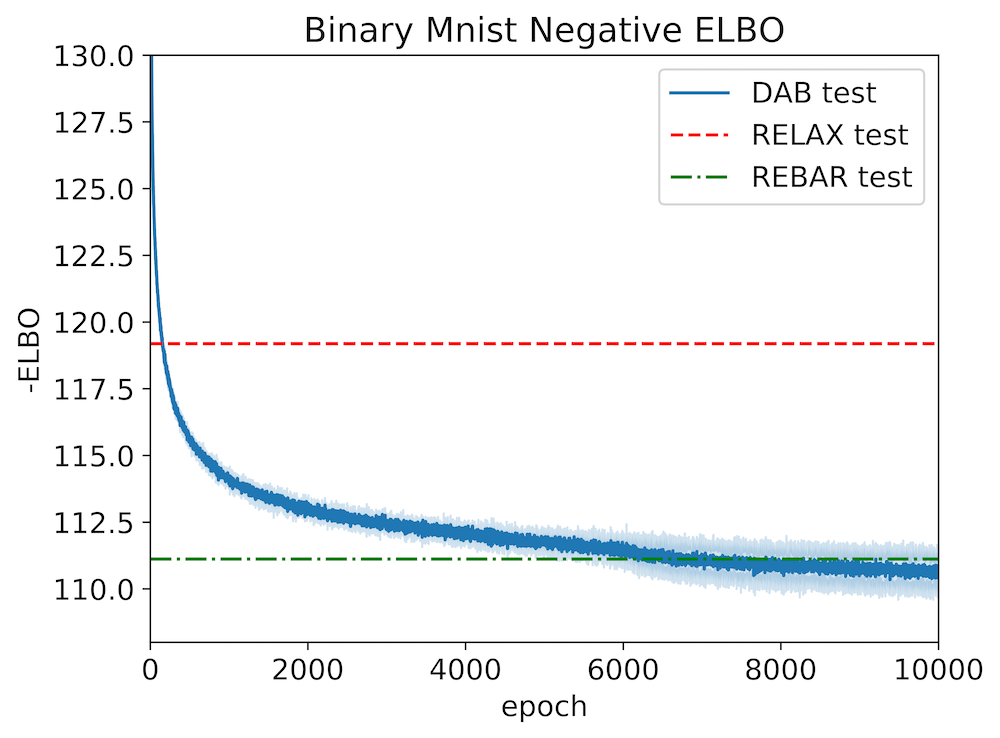}
  \end{minipage}
\begin{minipage}{0.5\textwidth}
  \includegraphics[width=\linewidth]{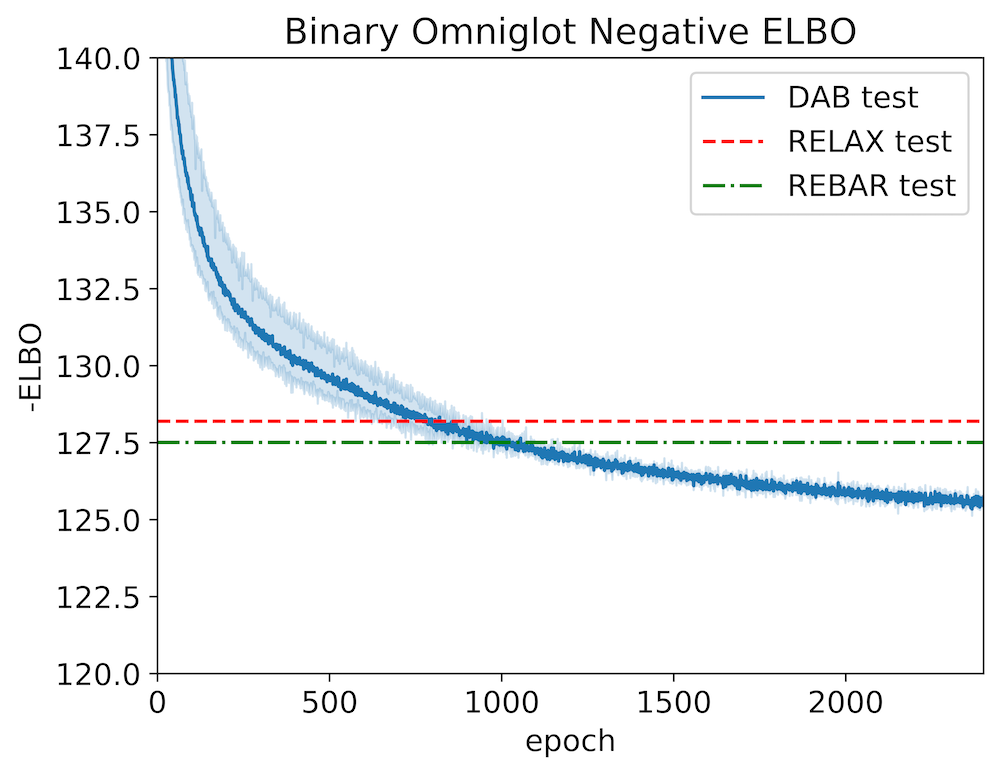}
  \end{minipage}
  \caption{Test Negative Variational Lower Bound (-ELBO) [lower being
    better]. \emph{Left}:  Binarized MNIST. \emph{Right}: Omniglot.}\label{train_test_mnist_omniglot}
\end{figure}%

\noindent While REBAR and RELAX used learning rate
schedulers, we used a fixed learning rate of 3e-4 with Adam for the entirety of
training. We believe that using a scheduled learning rate will improve
time to convergence for DAB, but leave this for future work.

\subsection{Imagenet Test Curves for MS-SSIM and Linear Separability}\label{imagenet_curves}
The full test curves for Imagenet are shown below:
\begin{figure}[H]
  % \vspace{-0.3in}
\begin{minipage}{0.5\textwidth}
    \includegraphics[width=\linewidth]{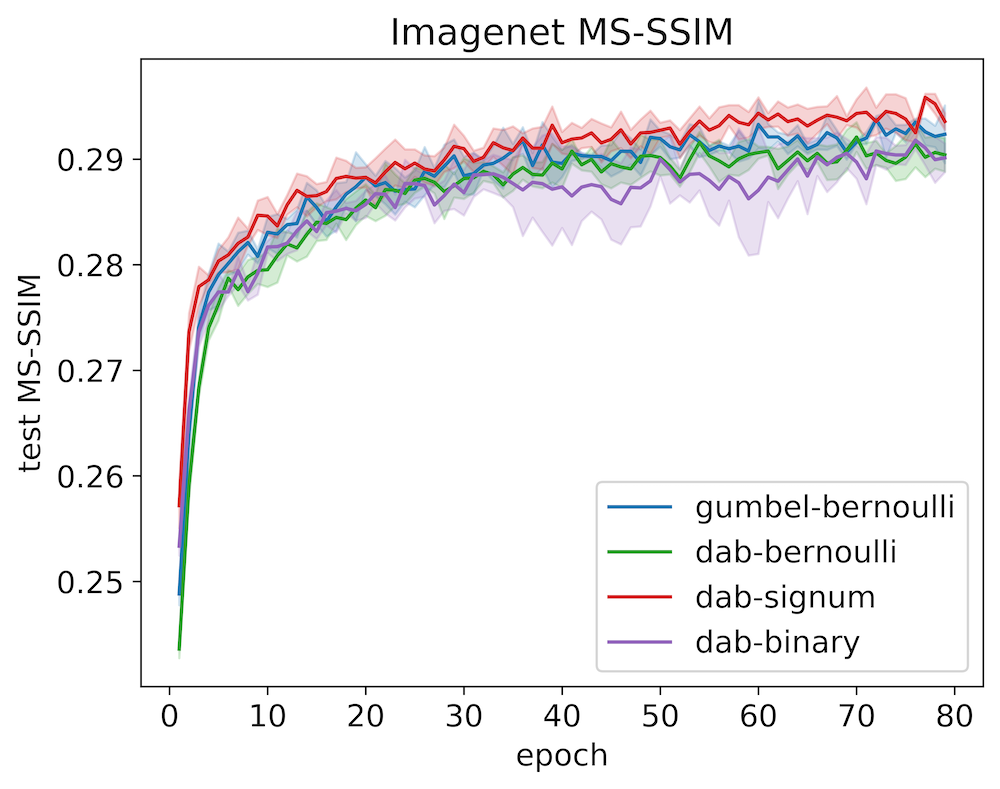}
\end{minipage}%
\begin{minipage}{0.5\textwidth}
  \includegraphics[width=\linewidth]{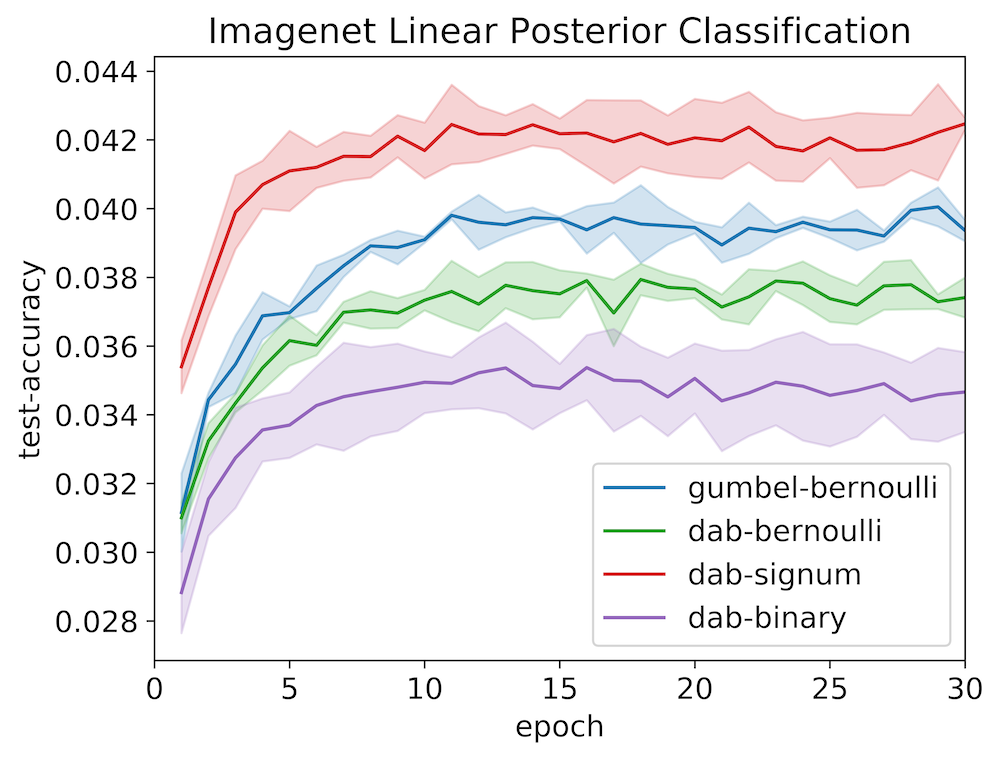}
\end{minipage}%
\caption{Five trials (each) of ImageNet using only BPP=0.00097 due to computational
restrictions (results reported as mean $\pm$ std). \emph{Left:} Test MS-SSIM \cite{wang2003multiscale}. \emph{Right:} \textbf{Purely
unsupervised} linear posterior test-classification accuracy; Images
are compressed from $\mathbb{R}^{512 \times 512}$ to \textbf{786} bits (\textbf{496} for
\emph{dab-signum} since $786 \approx 496 \log_2(3)$) and yield a \textbf{40x} improvement over random
guessing (0.001).}\label{imagenet_results_full}
\end{figure}

\subsection{Model Hyper-Parameters}
{\centering
  \begin{table}[H]
    \begin{center}
  \scalebox{0.8}{\parbox{1.0\linewidth}{%
{\renewcommand{\arraystretch}{1.3}%
\begin{tabular}{l|c|c|c|c|c|}
\cline{2-6}
                                             & \textbf{FashionMNIST} & \textbf{CIFAR10}                                                             & \textbf{ImageNet}                                                          & \textbf{Sorting}               & \textbf{Classification} \\ \hline
\multicolumn{1}{|l|}{\textbf{Optimizer}}     & Adam                  & RMSProp                                                                      & RMSProp                                                                    & Adam                           & Adam                    \\ \hline
\multicolumn{1}{|l|}{\textbf{LR}}            & 1e-3                  & 1e-4                                                                         & 1e-4                                                                       & 1e-4                           & 1e-4                    \\ \hline
\multicolumn{1}{|l|}{\textbf{Batch-Size}}    & 128                   & 128                                                                          & 192                                                                        & 1024                           & 128                     \\ \hline
\multicolumn{1}{|l|}{\textbf{Activation}}    & ELU                   & ReLU                                                                         & ELU                                                                        & Tanh                           & ELU                     \\ \hline
\multicolumn{1}{|l|}{\textbf{Normalization}} & Batchnorm             & \begin{tabular}[c]{@{}c@{}}Batchnorm-Conv,\\ None-Dense\end{tabular}        & \begin{tabular}[c]{@{}c@{}}Batchnorm-Conv,\\ None-Dense\end{tabular}      & None                           & Batchnorm               \\ \hline
\multicolumn{1}{|l|}{\textbf{Layer-Type}}    & Similar to U-Net     & \begin{tabular}[c]{@{}c@{}}Coord-Conv encoder,\\ Dense decoder\end{tabular} & \begin{tabular}[c]{@{}c@{}}Resnet18 encoder,\\ Dense decoder\end{tabular} & LSTM (gradclip 5) + Dense(256) & CifarResnet18           \\ \hline
\multicolumn{1}{|l|}{\textbf{DAB-$\gamma$}}     & 10                    & 70                                                                           & 2                                                                          & 10                             & 10                      \\ \hline
\end{tabular}
}}}
\end{center}
\end{table}
}

\subsection{Bayesian Interpretation of DAB}\label{bayesian_model_sec}

\begin{figure}[H]
  \begin{center}
    \includegraphics[width=100mm]{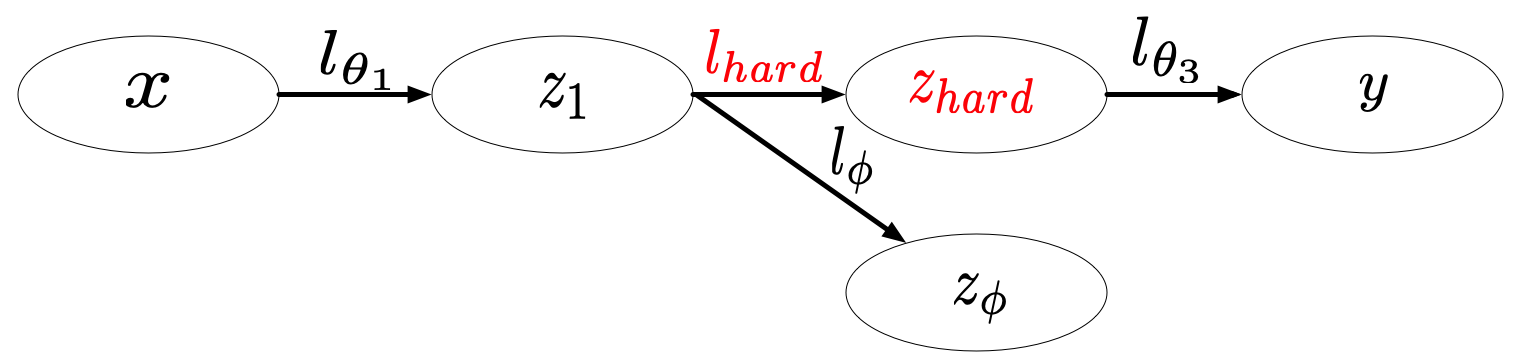}
\end{center}
\caption{Graphical model of our proposed framework. $\color{red}l_{hard}$ represents
  the non-differentiable function and $\color{red}z_{hard}$ its
  (latent variable) outputs.}\label{graphical_model}
\end{figure}%

A graphical model depicting a generic version of our framework is shown in Figure \ref{graphical_model_v2}. Given some true input data
distribution, $x \sim p(x), y \sim p(y|x)$, and a set of $J$ ($J=3$ in Figure \ref{graphical_model}) functional approximators,
$l_{\theta_i}: \mathbb{R}^N \mapsto \mathbb{R}^N,\ i \in \{1..J\}$, our
learning objective is defined as maximizing the log likelihood, $\log
p_{\theta}(y|x)$, coupled with a new regularizer, $\log p_{\phi}(z_{\text{hard}} | z_{\phi})$
($\mathcal{L}_{DAB}$ in Figure \ref{graphical_model}),
introduced in this work: % Our objective, $\max_{\theta, \phi}\
% \mathcal{J}(\theta, \phi)$, is:
% \vskip -0.2in
\begin{align}
   &\max_{\theta, \phi}\ \mathcal{J}(\theta, \phi)= \mathbb{E}_{x} [\ \log
                                                   p_{\theta}(y|x) +
                                                   \gamma \ \log
                                                   p_{\phi}(z_{\text{hard}} |
                                                   z_{\phi})\ ],\label{line1} \\
                                                 &= \mathbb{E}_{x} [\ \log
                                                   p_{\theta}(y|z_{\text{hard}})\ p_{\phi}(z_{\text{hard}}|z_1)\
                                                   p_{\theta}(z_1|x)
                                                   + \gamma \ \log
                                                   p_{\phi}(z_{\text{hard}} |
                                                   z_{\phi})\ ], \label{eqn4} \\
                                                 &= \mathbb{E}_{x} [\ \log
                                                   p_{\theta}(y|l_{\theta_3}({\color{red}l_{\text{hard}}}(l_{\theta_1}(x))))
                                                   + \gamma \ \log
                                                   p_{\phi}(z_{\text{hard}} |
                                                   l_{\phi}(z_1))\ ]. \label{main_eqn}
\end{align}

We transition from Equation \ref{line1} to Equation \ref{eqn4} by using the
conditional independence assumptions from our graphical model in
Figure \ref{graphical_model}. Since the latent representations $z_i$ are simple functional
transformations, we can represent the distributions $,p(z_i |
z_{i-1}),\ i > 0$\footnote{$z^0 := x$.} (Equation \ref{eqn4}), by dirac distributions centered around their
functional evaluations: $z_i | z_{i-1} \sim
\delta(l_{\theta_i}(z_{i-1}))$. This allows us to rewrite our
objective as shown in Equation \ref{main_eqn}, where $\gamma$ is a
problem specific hyper-parameter. A key point is that during the
forward functional evaluations of the model we
use the non-differentiable function, ${\color{red}l_{\text{hard}}}$.

Note that we have kept our framework generic. In the following section
we will describe the gaussianity assumptions on the regularizer
$p_{\phi}(z_{\text{hard}}| z_1)$ and why it converges.

% \vspace{-0.15in}
\subsubsection{Choice of metric under simplifying assumptions}\label{regularizer}

In this section we analyze the regularizer introduced in Equation
\ref{main_eqn} / Equation \ref{dab_loss} in the special case where the non-differentiable
function output, $z_{\text{hard}} = {\color{red}l_{\text{hard}}}(z_1) = \phi z_1 + \epsilon$,
is a (differentiable) linear transformation of the previous layer coupled with
additive Gaussian noise (aleatoric uncertainty):
% \vskip -0.2in
% \vspace{-0.15in}
\begin{align}
  z_{\text{hard}} &= \phi z_1 + \epsilon,\ \epsilon \sim \mathcal{N}(0, \sigma^2),  \\
  z_{\text{hard}} | \phi z_1, \sigma^2 &\sim \prod_{i=1}^N
                                           \mathcal{N}(\phi
                                           z_1,\sigma^2). \label{gauss_likelihood}
\end{align}
% \vskip -0.1in
Under these simplifying assumptions our model induces a Gaussian
log-likelihood as shown in Equation \ref{gauss_likelihood}.
At this point we can directly maximize the above likelihood using maximum
likelihood estimation. Alternatively, if we have apriori knowledge we can introduce
it as a prior, $p(\phi)$, over the weights $\phi$, and minimize the
negative log-likelihood multiplied by the prior to evaluate the
posterior (which will be induced as a gaussian),
i.e. the MAP estimate. If we make a conjugate prior assumption, $p(\phi) \sim
\mathcal{N}(0, \sigma^2_{\phi})$, then:
% \vspace{-0.2in}
\begin{align}
  -\log (\text{posterior}) &\propto - \log \prod_{i=1}^N
  \underbrace{\mathcal{N}(\phi z_1,\sigma^2)}_{\text{likelihood}}
  \underbrace{\mathcal{N}(0, \sigma^2_{\phi})}_{\text{prior}}, \\
  &= \sum_{i=1}^N \frac{-1}{\sigma^2} (z_{\text{hard}} - \phi z_1)^2 -
    \frac{\phi^2}{\sigma_{\phi}^2} + \text{const}, \\
  &\propto || z_{\text{hard}} - \phi z_1||_2^2. \label{metric}
\end{align}
% \vskip -0.1in
This analysis leads us to the well known result that a linear transformation
with aleatoric Gaussian noise results in a loss proportional to the L2
loss (Equation \ref{metric}). However, what can we say about the case where
$z_{\text{hard}}$ is a non-linear, non-differentiable output? In
practice we observe that using the L2 loss, coupled with a
non-linear neural network transformation, $l_{\phi}(z_1)$ produces
strong results. To understand why, we appeal to the central limit
theorem which states that the scaled mean of the random
variable converges to a Gaussian distribution as the sample size increases. Furthermore, if we can
assume a zero mean, positive variance, and finite absolute third
moment, it can be shown that the rate of convergence to a Gaussian distribution is
proportional to $\frac{1}{\sqrt{N}}$, where $N$ is the number of
samples \cite{berry1941accuracy}.
\\

\noindent{\textbf{Sketch of Convergence Proof}}:
\begin{table}[H]
  \begin{center}
\begin{tabular}{|l|}
  \hline
  \textbf{Given}: the true distribution converges to a Gaussian (in
  the limit) as per Section \ref{regularizer}. \\
  \textbf{Given}: neural networks are universal function approximators
  \cite{hornik1989multilayer}, \\\hspace{0.44in}then in the limit, $l_{\phi}(z_1) \approx
  l_{\text{hard}}(z_1)$ which implies $\frac{\delta
  l_{\phi}}{\delta l_{\theta_1}} \approx \frac{\delta
  l_{\text{hard}}}{\delta l_{\theta_1}}$. \\ \\
  Then DAB converges via the least-squares estimator to the true
  distribution via the least-square estimator \\convergence proofs of \cite{lai1982least,lai1991recursive,guo1996self,bittanti1990recursive}.
 \\ \hline
\end{tabular}
\end{center}
\end{table}

We empirically explored alternatives such as the
Huber loss \cite{huber1992robust}, cosine loss, L1 loss and cross-entropy loss, but found the
L2 loss to consistently produce strong results and we use it for all
presented experiments.

\end{document}